\documentclass[10pt,twocolumn,letterpaper]{article}
\usepackage[pagenumbers]{cvpr}

\usepackage{microtype}

\usepackage{multirow}
\usepackage{pgfplots}
\usepackage{stfloats}
\pgfplotsset{compat=1.18}
\usepgfplotslibrary{groupplots}

\AtEndPreamble{\AtBeginDocument{\crefname{appendix}{Supplementary}{Supplementary}\Crefname{appendix}{Supplementary}{Supplementary}}}

\definecolor{cvprblue}{rgb}{0.21,0.49,0.74}
\usepackage[pagebackref,breaklinks,colorlinks,allcolors=cvprblue]{hyperref}
\hypersetup{
    pdftitle={HiT: Human Inertial Transformer},
    pdfauthor={Yiquan Li, Jiawei Zhang, Chenfeng Gao, Karan Ahuja}
}

\title{HiT: Human Inertial Transformer}
\author{%
    Yiquan Li\textsuperscript{*}, Jiawei Zhang\textsuperscript{*}, Chenfeng Gao, Karan Ahuja\\
    Northwestern University\\
    {\small\ttfamily \{yiquanli2030, jiaweizhang2027, chenfenggao2029\}@u.northwestern.edu}\\
    {\small\ttfamily kahuja@northwestern.edu}\\[2pt]
    {\small \textsuperscript{*}Equal contribution}
}

\begin{document}
\maketitle
\begin{abstract}
Inertial sensors in glasses, watches, and phones observe human motion all day, yet existing models either estimate pose from a fixed, calibrated sensor setup or map the signals directly to activity labels and text without an explicit model of the body.
We present HiT (Human Inertial Transformer), a unified model that estimates full-body pose from any subset of body-worn IMUs and describes the observed motion in natural language without calibration or sensor placement knowledge. We first train HiT on real-world motion capture to estimate pose and the placement of every sensor, then freeze it and output language from it with a language head supervised jointly by activity labels, narration alignment, and captioning.
Trained only on Nymeria, HiT raises zero-shot activity recognition accuracy across ten unseen datasets from 43.9\% (AnyMo) to 65.1\%, improves IMU--text retrieval MRR 1.8- to 2.2-fold on the full Nymeria test pool and on zero-shot Ego-Exo4D, and raises captioning BLEU-4 from 6.5 to 11.0.
Without T-pose calibration and without knowing the sensor sites, its pose estimates are competitive with calibrated methods across datasets, more accurate than IMU4D zero-shot on IMUPoser and DIP-IMU; it also recognizes the body segment of 93.6\% of the sensors on unseen sequences. Code and model will be released.
\end{abstract}

\section{Introduction}
\label{sec:intro}

\begin{figure}[t]
\centering
\includegraphics[width=\linewidth]{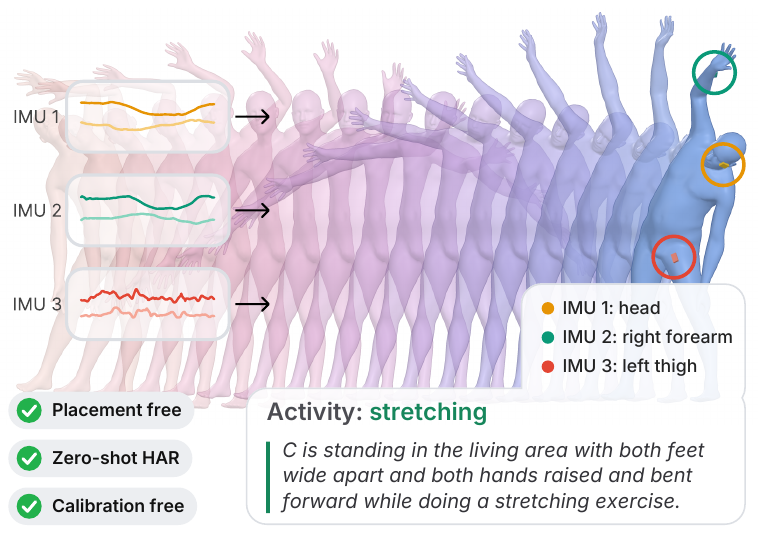}
\caption{HiT reads the raw signals of one to six body-worn IMUs without calibration and without being told where each device is worn.
From these signals one model estimates the full-body pose over time, the body segment of every sensor, the activity by zero-shot recognition, and a caption of the motion.}
\label{fig:teaser}
\end{figure}

\begin{figure*}[t]
\centering
\includegraphics[width=\textwidth]{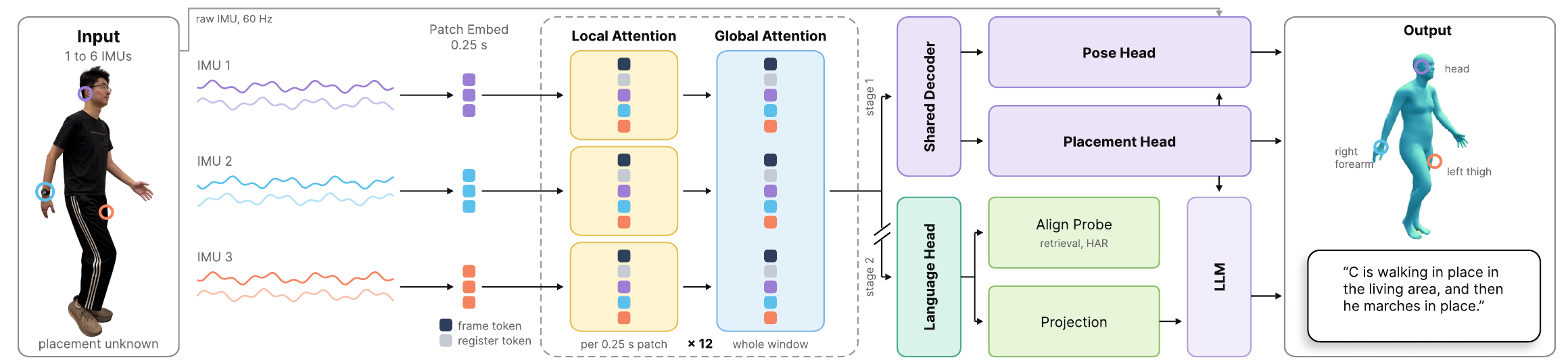}
\caption{Overview of HiT.
Stage~1: the raw signals of any subset of body-worn IMUs, uncalibrated and without a label of where each is worn, become patch tokens; a backbone alternates attention across sensors within a time step and attention over the whole window; a decoder with pose and sensor queries feeds a pose head (full-body pose at 60\,Hz) and a placement head (body segment, angle around the segment and orientation of every sensor).
Stage~2: the backbone and both heads are frozen, and a language head that reads the backbone's features over time segments is trained with narration alignment, activity labels and captioning.}
\label{fig:arch}
\end{figure*}

IMUs in the glasses, watches, earbuds, and phones people already wear can capture their movements throughout the day with minimal power consumption, even when cameras and microphones are impractical, making them well suited for understanding everyday human motion. Current inertial systems miss two requirements of real-world use.
First, they need calibration: pose estimators~\cite{dip,transpose,pip,imuposer,mobileposer,probip,imu4d} align every device to its body segment by a T-pose and to a shared world frame, which users of consumer devices do not do.
Second, pose estimators and activity and language models such as UniMTS~\cite{unimts} and AnyMo~\cite{anymo} are told where each sensor is worn, whereas a phone may sit in a pocket or a hand and a watch on either wrist.

We therefore rethink how an inertial system is built to tackle these challenges.
First, we avoid relying on calibrated sensor orientations. Instead, HiT uses raw angular velocity and acceleration to capture the motion of each body segment. By training with randomized mounting rotations, it learns to estimate pose despite changes in how a device is oriented on the body.
Second, inertial signals capture the local dynamics of the body part that carries the device, and all devices move with one body, so a model of human motion can map these dynamics onto the body and place each sensor itself, achieving pose estimation without placement knowledge.

Following these ideas, we propose HiT, the Human Inertial Transformer (\cref{fig:arch}), which, trained on the Nymeria dataset~\cite{nymeria}, predicts the full-body pose and the placement of every sensor from any subset of body-worn IMUs.
We further connect it to activity recognition and language: instead of mapping the signals directly to words~\cite{imu2clip,unimts,anymo,sensorllm}, HiT outputs language from the frozen pose backbone, supervised by activity labels, narration alignment and captioning, so that the language is grounded in a body-pose representation.
Our contributions are:

(1) a calibration-free and placement-free IMU-to-pose backbone for one to six sensors on any of 17 body segments, whose pose is more accurate than IMU4D's zero-shot on IMUPoser and DIP-IMU and unchanged when every sensor is remounted, and whose placement head recognizes the body segment of 93.6\% of the sensors;

(2) language output from the frozen pose backbone, which with the same narrations beats end-to-end language training and, trained only on Nymeria, raises zero-shot activity recognition on ten unseen datasets from 43.9\% (AnyMo) to 65.1\% and improves retrieval and captioning over prior IMU--language models.

\section{Related Work}
\label{sec:related}

\paragraph{Human Pose from sparse IMUs.}

Full-body pose estimation from a few inertial sensors has drawn considerable attention in recent years, with early works such as SIP~\cite{sip} and DIP~\cite{dip}. Subsequent methods, including TransPose~\cite{transpose}, PIP~\cite{pip}, TIP~\cite{tip}, DynaIP~\cite{dynaip}, PNP~\cite{pnp}, and ProbIP~\cite{probip}, introduced translation estimation, physics, uncertainty modeling, and improved sequence models. IMUPoser~\cite{imuposer} and MobilePoser~\cite{mobileposer} extended the task to subsets of commercial devices; WHIP~\cite{whip} adds a head-mounted device and pressure insoles, while IMU4D~\cite{imu4d} interprets such signals using a large language model and incorporates object interactions. These methods require sensor orientations from the devices' fusion filters to be aligned with the bones using a T-pose and expressed in a common world frame. In our experiments, the evaluated calibration-dependent baselines exhibit approximately two- to three-fold increases in pose error under simulated remounting (\cref{sec:exp}). Some works have attempted to address the calibration problem. TIC~\cite{tic} corrects mounting offsets online but is evaluated after an initial static calibration, while ArmPoser~\cite{armposer} estimates arm pose from a single uncalibrated watch rather than full-body pose; Loose Inertial Poser~\cite{lip} models IMUs that move with a loose jacket.
 In addition to these calibration requirements, the aforementioned methods require sensor placement information to be explicitly provided as input, limiting their applicability when sensor placement is unknown; DiffusionPoser~\cite{diffusionposer} accepts arbitrary sensor combinations with one diffusion model, but at given placements.
 Recognizing where a device is worn from its accelerometer signal has previously been treated as a separate classification task~\cite{kunze2005whereami,vahdatpour2011onbody}, but these studies do not address whether full-body pose can be predicted without knowing sensor placement. Beyond body-worn IMUs, full-body motion is also reconstructed from the sparse head and hand signals of XR headsets~\cite{avatarposer,egoego,hmdposer,egoposer,flowavatar}.

\paragraph{Activity and language from IMUs.}
Inertial activity recognition has moved from supervised classifiers over fixed categories, such as DeepConvLSTM~\cite{deepconvlstm}, to representations that transfer to new activities and sensor setups: TransfHAR~\cite{transfhar} adapts pretrained wrist features from a few labeled examples, and UniMTS~\cite{unimts} and AnyMo~\cite{anymo} recognize unseen activities from their names after training on IMUs simulated from motion capture at many body placements; self-supervised pre-training on large unlabeled wearable data~\cite{yuan2024ssl} also transfers across activities, and orientation-invariant transformations~\cite{yurtman2017} remove the device's orientation before recognition.
Language extends this from labels to retrieval and description: IMU2CLIP~\cite{imu2clip} and ImageBind~\cite{imagebind} align IMUs with text, directly or through video, by contrastive learning; SensorLLM~\cite{sensorllm} and LLaSA~\cite{llasa} condition language models on sensor features; AnyMo couples alignment with captioning, and IMU4D~\cite{imu4d} predicts body motion and descriptions together from calibrated orientations.
These models map the signals to words without first learning the body, and UniMTS and AnyMo are told where each sensor is worn. Closest to our goal, EgoLM~\cite{egolm}, Ego4o~\cite{ego4o} and Mojito~\cite{mojito} track motion and generate language in one model, but from egocentric video or from sensors at known placements.
HiT learns body pose and sensor placement first and outputs language from the frozen backbone, which with the same narrations improves zero-shot recognition, retrieval and captioning over end-to-end language training (\cref{sec:exp:posesup}).

\section{Method}
\label{sec:method}

To tackle the problems of full-body pose estimation, action narration alignment, caption generation, and human activity recognition (HAR) from inertial sensors anywhere on the human body, we introduce Human Inertial Transformer (HiT). HiT builds language supervision on a backbone pretrained for pose estimation and sensor placement, so that the language-related tasks learn from a pose-grounded representation. We start by introducing the problem in Sec.~\ref{sec:method:problem} and the sensor augmentation in Sec.~\ref{sec:method:input}, followed by our backbone architecture in Sec.~\ref{sec:method:backbone} and the corresponding language head in Sec.~\ref{sec:method:lang}. Hyperparameters are listed in Supplementary~G.

\subsection{Problem formulation}
\label{sec:method:problem}

A person wears $S$ devices ($1 \le S \le 6$), each on one of 17 body segments (pelvis, chest, head, and the shoulders, upper arms, forearms, hands, thighs, shanks and feet of both sides).
Device $i$ records its specific force and angular velocity $\mathbf{x}_i \in \mathbb{R}^{T \times 6}$ at 60\,Hz in its own frame.
Its placement $\mathbf{p}_i = (s_i, \theta_i, \mathbf{R}_i)$ is unknown: the segment $s_i$, the angle $\theta_i$ of its skin normal around the segment's long axis, and its orientation $\mathbf{R}_i$.
We assume only that one device axis is approximately the outward skin normal (a watch face, a phone screen, the front of glasses), so the unknown mounting is a spin about this normal, a small tilt and possibly a flip.
From $\{\mathbf{x}_i\}$ alone, HiT estimates (i) the full-body pose at every frame (SMPL joint rotations, root velocity, yaw rate, pelvis height, foot contacts); (ii) the placement $\hat{\mathbf{p}}_i$ of every device; and (iii) for every time segment, an embedding in a text space and a caption.
Pose estimators instead obtain the orientations from a fusion filter, a T-pose and a common world heading, and are given $s_i$ as input~\cite{dip,transpose,mobileposer,imu4d}; in HiT both are outputs.
Being calibration-free and estimating the placement both make HiT deployable in the real world, where users neither stand in a T-pose nor report where each device is worn.
Estimating the placement also makes the backbone keep track of which signal comes from which sensor, which the pose head and the caption prompt rely on (\cref{sec:method:backbone}).

\subsection{Sensor augmentation}
\label{sec:method:input}

\begin{figure}[t]
\centering
\includegraphics[width=0.74\linewidth]{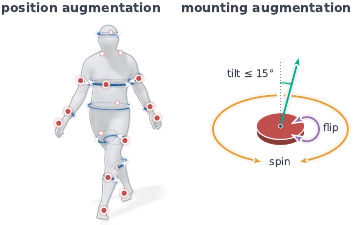}
\caption{Sensor augmentation in training.
Left, position augmentation: the sensors of the motion-capture suit (red; open: behind the body) are each moved around their body segment along the blue ring, most often by a small angle (the ring fades where they are moved less often); the sensors on the shoulders, hands and feet are not moved.
Right, mounting augmentation, in place: the device is spun about the skin normal, tilted by up to $15^\circ$ and flipped.}
\label{fig:aug}
\end{figure}

\paragraph{Mounting.}
During training every sensor receives a random spin about the skin normal, a tilt and a flip (\cref{fig:aug}), following AnyMo~\cite{anymo}, as well as gain, bias, noise, lower sampling rates and gyroscope dropout, and each window keeps a random subset of the sensors.
As input normalization, every sensor is also rotated about the skin normal so that the in-plane component of its mean acceleration, which is dominated by gravity, points along $+x$; the angle between gravity and the skin normal, which tells, e.g., whether a thigh is upright (standing) or horizontal (sitting), is unchanged.

\paragraph{Position.}
The sensors of the motion-capture suit used for training (\cref{sec:exp:setup}) sit at one fixed place on their segment, whereas a watch or a phone can sit anywhere around a limb, so in training each sensor is moved around the long axis of its segment by a random angle (position augmentation, \cref{fig:aug}; 40\% stay, 40\% move by up to $90^\circ$, 20\% anywhere): its frame is turned and its specific force receives the lever-arm terms $\boldsymbol{\alpha}\times\Delta\mathbf{r} + \boldsymbol{\omega}\times(\boldsymbol{\omega}\times\Delta\mathbf{r})$ of the displacement $\Delta\mathbf{r}$, which is exact for a rigid segment.

\begin{table*}[b]
\caption{Pose estimation: MPJPE and PA-MPJPE in cm, MPJRE in degrees; best in bold, second best underlined.
HiT (ours) is the model of all other tables.
Calibration-free: the method reads the raw signals of each device; placement-free: the method is not told where each sensor is worn.
Grey: ProbIP's released model takes only the wrists and the head, so it uses both watches and the earbuds on IMUPoser and 3 of the 5 IMUs on DIP-IMU; these cells are not ranked.
The pre-training data of MobilePoser and IMU4D likely contain TotalCapture's motions.
Settings in Supplementary~A.}
\label{tab:pose}
\centering
\small
\setlength{\tabcolsep}{3.0pt}
\resizebox{\textwidth}{!}{%
\begin{tabular}{lcc|ccc|ccc|ccc}
\toprule
& & & \multicolumn{3}{c|}{IMUPoser} & \multicolumn{3}{c|}{DIP-IMU (5 IMUs)} & \multicolumn{3}{c}{TotalCapture (3 IMUs)} \\
Method & \begin{tabular}{@{}c@{}}Calibration-\\free\end{tabular} & \begin{tabular}{@{}c@{}}Placement-\\free\end{tabular} & MPJPE$\downarrow$ & MPJRE$\downarrow$ & PA-MPJPE$\downarrow$ & MPJPE$\downarrow$ & MPJRE$\downarrow$ & PA-MPJPE$\downarrow$ & MPJPE$\downarrow$ & MPJRE$\downarrow$ & PA-MPJPE$\downarrow$ \\
\midrule
\multicolumn{12}{l}{\emph{Zero-shot (not trained on the target dataset)}} \\
MobilePoser & $\times$ & $\times$ & -- & -- & -- & -- & -- & -- & 10.22 & 22.7 & 7.67 \\
ProbIP & $\times$ & $\times$ & {\color{gray}9.11} & {\color{gray}20.9} & {\color{gray}7.17} & -- & -- & -- & \underline{9.72} & \underline{21.3} & \underline{6.71} \\
IMU4D & $\times$ & $\times$ & \underline{15.03} & \underline{29.7} & \underline{12.58} & \underline{8.78} & \underline{25.1} & \underline{6.24} & \textbf{9.68} & \textbf{19.7} & 7.35 \\
HiT (ours) & \checkmark & \checkmark & \textbf{11.24} & \textbf{28.4} & \textbf{8.93} & \textbf{7.82} & \textbf{22.4} & \textbf{6.08} & 10.43 & 25.6 & \textbf{6.46} \\
\midrule
\multicolumn{12}{l}{\emph{Trained / Fine-tuned on the target dataset}} \\
MobilePoser & $\times$ & $\times$ & 8.15 & \underline{19.6} & 7.05 & 8.79 & 25.1 & 7.69 & 9.14 & 18.6 & 7.12 \\
ProbIP & $\times$ & $\times$ & {\color{gray}6.86} & {\color{gray}15.6} & {\color{gray}5.39} & {\color{gray}8.07} & {\color{gray}24.4} & {\color{gray}5.96} & 9.72 & 19.3 & 7.33 \\
IMU4D & $\times$ & $\times$ & \underline{7.66} & 21.7 & \underline{6.72} & \textbf{5.11} & \textbf{11.5} & \textbf{4.01} & \textbf{7.42} & \textbf{15.3} & \textbf{5.69} \\
HiT (ours) & \checkmark & \checkmark & \textbf{6.29} & \textbf{16.3} & \textbf{5.48} & \underline{5.30} & \underline{11.8} & \underline{4.22} & \underline{7.91} & \underline{16.4} & \underline{5.77} \\
\bottomrule
\end{tabular}}

\end{table*}

\begin{figure*}[b]
\centering
\begin{subfigure}[t]{0.49\textwidth}
\centering
\includegraphics[width=\linewidth]{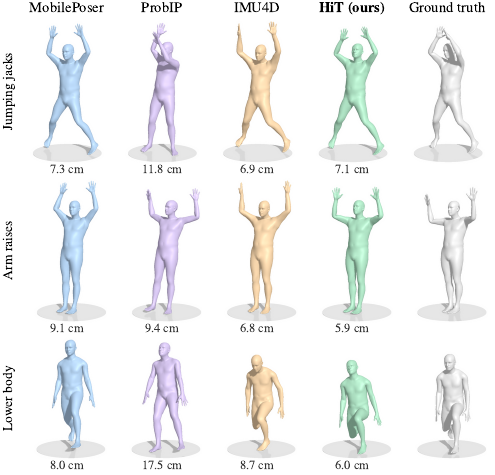}
\caption{Calibrated inputs}
\label{fig:qual_cal}
\end{subfigure}\hfill
\begin{subfigure}[t]{0.49\textwidth}
\centering
\includegraphics[width=\linewidth]{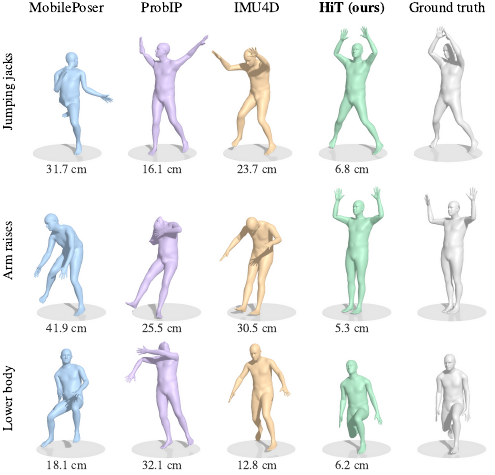}
\caption{Inputs without pose calibration}
\label{fig:qual_uncal}
\end{subfigure}
\caption{Pose on IMUPoser (P9--P10) with calibrated inputs (a) and with a random mounting rotation of every device (b).
MobilePoser, IMU4D and HiT read the left watch, the right-pocket phone and the earbuds and are trained on P1--P8; ProbIP reads both watches and the earbuds and starts from the ground-truth pose.
In (b), every device is given a random mounting rotation (up to $180^\circ$ about the skin normal, $15^\circ$ of tilt) and a random world heading.
Numbers: MPJPE of the frame.
The calibrated baselines break without calibration; HiT's pose is unchanged.}
\label{fig:qual}
\end{figure*}

\subsection{Backbone and prediction heads}
\label{sec:method:backbone}

Each 0.25 s patch of each sensor (15 samples of ten channels: the three axes of the normalized specific force and of the angular velocity, and the full and in-plane norm of each) is embedded by an MLP into one token; all sensors share one embedding, so the tokens carry no site label.
The backbone follows VGGT~\cite{vggt}, a time patch playing the role of an image and its sensors that of the image's patches: per patch we add a frame token and registers~\cite{registers}, and \emph{local} blocks within one patch alternate with \emph{global} blocks over all tokens of the window, with rotary position embeddings~\cite{rope}. A shared decoder holds, per patch, five pose queries, one per body part (torso, two arms, two legs), and one sensor query per sensor; all queries read the backbone's tokens at four depths. The decoded pose queries go to the pose head, and the decoded sensor queries to the placement head (Supplementary~F).

\paragraph{Placement head.}
From each sensor query, the placement head predicts the sensor's body segment (17 classes), the angle of the device around the segment's long axis as a von Mises distribution, and the device orientation in the body's heading frame at every patch with a Laplace scale; it is trained on the labels that the suit and the position augmentation provide. The head also pools each sensor's queries over the window into one vector, its descriptor, which the pose head and the caption prompt use to tell the sensors apart.

\paragraph{Pose head.}
A DPT-style temporal fusion~\cite{dpt} fuses the decoded pose queries at 1, 2 and 4\,Hz and upsamples them to 60\,Hz, with a skip connection from the raw samples that is routed to each body part by attention over the sensors' descriptors, since no site label says which sensor belongs to which part; each part writes the outputs of its own SMPL~\cite{smpl} joints (\cref{sec:method:problem}), the torso also those of the root and each leg its foot contact.
Without a world frame, heading is estimated as a rate rather than an absolute value.
Stage~1 minimizes
\setlength{\abovedisplayskip}{3pt}\setlength{\belowdisplayskip}{3pt}%
\begin{equation}
\mathcal{L}_1 = \textstyle\sum_k w_k \mathcal{L}^{\text{pose}}_k + \lambda_{\text{p}} \left(\mathcal{L}_{\text{seg}} + \mathcal{L}_{\text{ang}} + \mathcal{L}_{\text{ori}}\right),
\label{eq:stage1}
\end{equation}
over the pose errors and the placement negative log-likelihoods (weights in Supplementary~F).

\subsection{Language output}
\label{sec:method:lang}

Stage~2 freezes all of stage~1 and trains one output head over the backbone's sensor and frame tokens at the same four depths.
Training windows are one to six consecutive time segments, each covering one of Nymeria's time-stamped narrations.
For every segment, six queries at its center read the keys inside the segment, and two window-level context queries read every key; all queries share one self-attention, so a segment token also sees the whole window.
A segment's tokens are pooled by learned attention probes~\cite{settransformer,scalingvit} and projected into the shared space of the frozen bge-large text encoder~\cite{bge} that the corresponding text is encoded with.

Three losses supervise the same segment tokens at three granularities:
\setlength{\abovedisplayskip}{3pt}\setlength{\belowdisplayskip}{3pt}%
\begin{equation}
\mathcal{L}_2 = \lambda_{\text{a}} \mathcal{L}_{\text{align}} + \lambda_{\text{l}} \mathcal{L}_{\text{label}} + \mathcal{L}_{\text{cap}}, \quad \lambda_{\text{a}} = \lambda_{\text{l}} = 0.5.
\label{eq:stage2}
\end{equation}
\emph{Narration alignment} $\mathcal{L}_{\text{align}}$: a SigLIP loss~\cite{siglip} between the segment embeddings and the bge-large embeddings of their narrations.
\emph{Activity labels} $\mathcal{L}_{\text{label}}$: a segment that coincides with a 5\,s window labeled by AnyMo-Bench~\cite{anymo} (IoU${}>0.9$) is scored in the same space against the embeddings of its 150 fine-grained labels (one-versus-rest sigmoid).
\emph{Captioning} $\mathcal{L}_{\text{cap}}$: the segment and context tokens, preceded by one token per sensor from the placement head's descriptor, are given as a soft prefix to Qwen2.5-0.5B~\cite{qwen25} with LoRA~\cite{lora}, trained to generate the segment's narration.
Supplementary~D shows that each loss serves its own output and that combining them costs none of the three.

\paragraph{Inference.}
\label{sec:method:inference}
For zero-shot activity recognition, a test window is one segment, and its alignment vector is compared with the projected bge-large embeddings of the dataset's class names; for retrieval, with the embeddings of the candidate sentences; for captioning, the LLM decodes from the segment and context tokens.

\section{Experiments}
\label{sec:exp}

\subsection{Setup}
\label{sec:exp:setup}

\paragraph{Training data.}
HiT is trained only on Nymeria~\cite{nymeria}: 813 sequences (212.7\,h) for pose and placement, and the 569 of them that carry narrations (113k narrations, 141.9\,h) for language. The training signals are real, from the 17 XSens sensors of its motion-capture suit and the IMUs of its glasses and wristbands (Supplementary~G). We report the language results of HiT with an average over three seeds of the language head on the same backbone.

\paragraph{Evaluation.}
We evaluate HiT performance on the three tasks it's designed to complete: full-body pose reconstruction, zero-shot and fine-tuned HAR, and finally caption generation. Pose wise, we evaluate HiT  with IMUPoser~\cite{imuposer} (MobilePoser's~\cite{mobileposer} protocol and evaluator), DIP-IMU~\cite{dip} and TotalCapture~\cite{totalcapture} (IMU4D's~\cite{imu4d} protocol); WHIP~\cite{whip} and the Nymeria validation split are in Supplementary~A.
We use the ten benchmarks of UniMTS~\cite{unimts} and AnyMo~\cite{anymo}, Opportunity~\cite{opportunity}, UCI-HAR~\cite{ucihar}, w-HAR~\cite{whar}, RealWorld~\cite{realworld}, TNDA-HAR~\cite{tndahar}, PAMAP2~\cite{pamap2}, USC-HAD~\cite{uschad}, WISDM~\cite{wisdm}, DSADS~\cite{dsads} and UTD-MHAD~\cite{utdmhad}, to evaluate the zero-shot HAR performance for HiT. The class names are extended to narrations with a fixed template (Supplementary~E).
On the caption generation task, we follow the Nymeria held-out protocol of AnyMo (3{,}908 windows from 20 unseen sequences; glasses and both wristbands) and zero-shot Ego-Exo4D~\cite{egoexo4d} (glasses only) to evaluate the language performance of HiT.

\subsection{Zero-shot and fine-tuned activity recognition}
\label{sec:exp:har}

\paragraph{Results.}
HiT raises the mean accuracy from 43.9\% (AnyMo~\cite{anymo}) to 65.1\% and macro-F1 from 36.6\% to 58.8\% (\cref{tab:har}), and is the most accurate on every dataset.
Accuracy is highest where the classes are postures and locomotion that Nymeria's narrations cover (TNDA-HAR, 99.8\%) and lowest on UTD-MHAD (24.5\%), most of whose 27 classes are arm gestures such as ``draw circle'' that the narrations do not describe; fine-tuning brings UTD-MHAD to 88.8\%, so the signals separate these classes and the zero-shot limit is on the text side.
At test time the model is not told where a benchmark's sensors are worn.
With full-shot fine-tuning the same model reaches 92.5\% (UniMTS 88.0\%).

\subsection{Language: retrieval and captioning}
\label{sec:exp:lang}
\begin{table*}[b]
\caption{Activity recognition on ten datasets (\%); the last three columns are means.
Top: zero-shot, none of the datasets seen in training.
Bottom: full-shot fine-tuning on UniMTS's training splits (implementation details in Supplementary~E).}
\label{tab:har}
\centering
\small
\begin{tabular}{lcccccccccc|ccc}
\toprule
& Opp. & UCI & w-HAR & RW & TNDA & PAMAP2 & USC & WISDM & DSADS & UTD & Acc & F1 & R@2 \\
classes & 4 & 6 & 7 & 8 & 8 & 12 & 12 & 18 & 19 & 27 & & & \\
\midrule
Chance & 25.0 & 16.7 & 14.3 & 12.5 & 12.5 & 8.3 & 8.3 & 5.6 & 5.3 & 3.7 & 11.2 & -- & 22.4 \\
\midrule
\multicolumn{14}{l}{\emph{Zero-shot}} \\
UniMTS & 45.9 & 35.2 & 59.0 & 43.6 & 59.1 & 47.2 & 30.5 & 27.8 & 31.5 & 22.8 & 40.3 & 33.7 & 57.0 \\
AnyMo & 59.4 & 56.5 & 57.4 & 48.4 & 59.4 & 52.6 & 27.7 & 25.4 & 36.3 & 16.3 & 43.9 & 36.6 & 69.7 \\
HiT (ours) & \textbf{73.6} & \textbf{66.6} & \textbf{76.0} & \textbf{81.8} & \textbf{99.8} & \textbf{74.0} & \textbf{54.6} & \textbf{48.1} & \textbf{51.7} & \textbf{24.5} & \textbf{65.1} & \textbf{58.8} & \textbf{80.0} \\
\midrule
\multicolumn{14}{l}{\emph{Full-shot fine-tuning}} \\
UniMTS & 88.2 & 92.1 & \textbf{98.4} & 84.0 & 99.6 & 98.0 & 58.3 & 81.3 & \textbf{94.3} & 85.6 & 88.0 & 88.0 & -- \\
HiT (ours) & \textbf{90.1} & \textbf{97.6} & 86.9 & \textbf{96.4} & \textbf{100.0} & \textbf{99.3} & \textbf{89.4} & \textbf{84.7} & 91.6 & \textbf{88.8} & \textbf{92.5} & \textbf{92.0} & -- \\
\bottomrule
\end{tabular}
\end{table*}

\begin{figure*}[b]
\centering
\includegraphics[width=\textwidth]{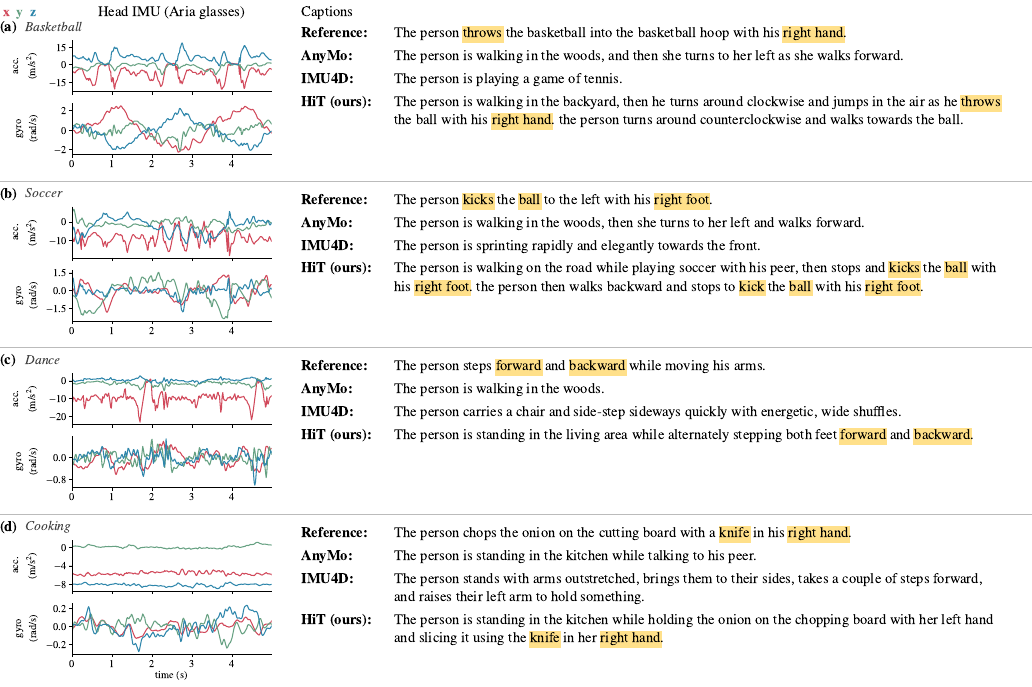}
\caption{Zero-shot captions on Ego-Exo4D.
Left: the head IMU (specific force and angular velocity, 60\,Hz).
Right: the reference and the captions of AnyMo (its released model, run by us), IMU4D and HiT.
Shaded: what HiT's caption shares with the reference.}
\label{fig:captions}
\end{figure*}

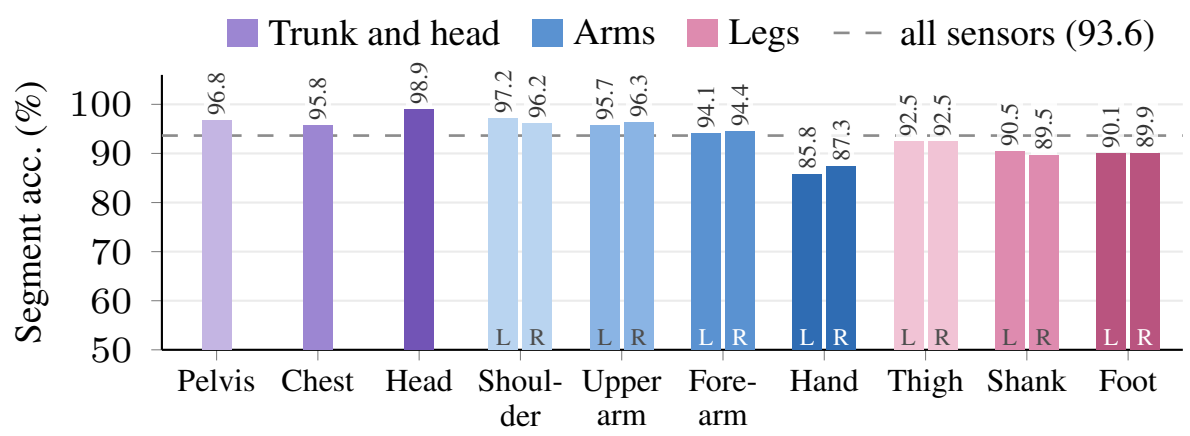
\begin{figure}[t]
\centering
\begin{tikzpicture}
\begin{axis}[
  ybar, bar width=5.6pt, width=\linewidth, height=3.4cm, scale only axis=false,
  ymin=50, ymax=106, ytick={50,60,70,80,90,100},
  ylabel={Segment acc.\ (\%)}, ylabel style={font=\scriptsize, yshift=-3pt},
  yticklabel style={font=\scriptsize},
  symbolic x coords={Pelvis,Chest,Head,Shoulder,Upper arm,Forearm,Hand,Thigh,Shank,Foot},
  xtick={Pelvis,Chest,Head,Shoulder,Upper arm,Forearm,Hand,Thigh,Shank,Foot},
  xticklabels={Pelvis,Chest,Head,Shoul-\\der,Upper\\arm,Fore-\\arm,Hand,Thigh,Shank,Foot},
  xticklabel style={font=\fontsize{6}{6.5}\selectfont, align=center, inner sep=1.5pt},
  enlarge x limits=0.06,
  axis x line*=bottom, axis y line*=left, tick align=outside, major tick length=2pt,
  ymajorgrids, grid style={black!8},
  point meta=explicit symbolic,
  nodes near coords, nodes near coords style={font=\fontsize{4.6}{4.6}\selectfont, color=black!80, rotate=90, anchor=west, inner sep=0.8pt, fill=white, fill opacity=0.85, text opacity=1},
  legend cell align=left, legend style={at={(1,1.0)}, anchor=south east, legend columns=4, font=\scriptsize, draw=none, fill=none, /tikz/every even column/.append style={column sep=4pt}},
  clip=false,
]
\draw[black!45, dashed, line width=0.5pt] (rel axis cs:0,0.7793) -- (rel axis cs:1,0.7793);
\addplot[fill={rgb,255:red,196;green,181;blue,227}, draw=none, bar shift=0pt, forget plot] coordinates {(Pelvis,96.77) [96.8]};
\addplot[fill={rgb,255:red,156;green,134;blue,210}, draw=none, bar shift=0pt, forget plot] coordinates {(Chest,95.79) [95.8]};
\addplot[fill={rgb,255:red,114;green,84;blue,182}, draw=none, bar shift=0pt, forget plot] coordinates {(Head,98.91) [98.9]};
\addplot[fill={rgb,255:red,185;green,212;blue,240}, draw=none, bar shift=-3.2pt, forget plot] coordinates {(Shoulder,97.16) [97.2]};
\node[font=\fontsize{4.5}{4.5}\selectfont, color=black!70, anchor=south, inner sep=0.8pt, xshift=-3.2pt] at (axis cs:Shoulder,50) {L};
\addplot[fill={rgb,255:red,185;green,212;blue,240}, draw=none, bar shift=3.2pt, forget plot] coordinates {(Shoulder,96.18) [96.2]};
\node[font=\fontsize{4.5}{4.5}\selectfont, color=black!70, anchor=south, inner sep=0.8pt, xshift=3.2pt] at (axis cs:Shoulder,50) {R};
\addplot[fill={rgb,255:red,138;green,180;blue,228}, draw=none, bar shift=-3.2pt, forget plot] coordinates {(Upper arm,95.66) [95.7]};
\node[font=\fontsize{4.5}{4.5}\selectfont, color=black!70, anchor=south, inner sep=0.8pt, xshift=-3.2pt] at (axis cs:Upper arm,50) {L};
\addplot[fill={rgb,255:red,138;green,180;blue,228}, draw=none, bar shift=3.2pt, forget plot] coordinates {(Upper arm,96.35) [96.3]};
\node[font=\fontsize{4.5}{4.5}\selectfont, color=black!70, anchor=south, inner sep=0.8pt, xshift=3.2pt] at (axis cs:Upper arm,50) {R};
\addplot[fill={rgb,255:red,90;green,146;blue,209}, draw=none, bar shift=-3.2pt, forget plot] coordinates {(Forearm,94.13) [94.1]};
\node[font=\fontsize{4.5}{4.5}\selectfont, color=white, anchor=south, inner sep=0.8pt, xshift=-3.2pt] at (axis cs:Forearm,50) {L};
\addplot[fill={rgb,255:red,90;green,146;blue,209}, draw=none, bar shift=3.2pt, forget plot] coordinates {(Forearm,94.41) [94.4]};
\node[font=\fontsize{4.5}{4.5}\selectfont, color=white, anchor=south, inner sep=0.8pt, xshift=3.2pt] at (axis cs:Forearm,50) {R};
\addplot[fill={rgb,255:red,47;green,108;blue,179}, draw=none, bar shift=-3.2pt, forget plot] coordinates {(Hand,85.77) [85.8]};
\node[font=\fontsize{4.5}{4.5}\selectfont, color=white, anchor=south, inner sep=0.8pt, xshift=-3.2pt] at (axis cs:Hand,50) {L};
\addplot[fill={rgb,255:red,47;green,108;blue,179}, draw=none, bar shift=3.2pt, forget plot] coordinates {(Hand,87.33) [87.3]};
\node[font=\fontsize{4.5}{4.5}\selectfont, color=white, anchor=south, inner sep=0.8pt, xshift=3.2pt] at (axis cs:Hand,50) {R};
\addplot[fill={rgb,255:red,241;green,195;blue,213}, draw=none, bar shift=-3.2pt, forget plot] coordinates {(Thigh,92.46) [92.5]};
\node[font=\fontsize{4.5}{4.5}\selectfont, color=black!70, anchor=south, inner sep=0.8pt, xshift=-3.2pt] at (axis cs:Thigh,50) {L};
\addplot[fill={rgb,255:red,241;green,195;blue,213}, draw=none, bar shift=3.2pt, forget plot] coordinates {(Thigh,92.45) [92.5]};
\node[font=\fontsize{4.5}{4.5}\selectfont, color=black!70, anchor=south, inner sep=0.8pt, xshift=3.2pt] at (axis cs:Thigh,50) {R};
\addplot[fill={rgb,255:red,222;green,139;blue,175}, draw=none, bar shift=-3.2pt, forget plot] coordinates {(Shank,90.45) [90.5]};
\node[font=\fontsize{4.5}{4.5}\selectfont, color=black!70, anchor=south, inner sep=0.8pt, xshift=-3.2pt] at (axis cs:Shank,50) {L};
\addplot[fill={rgb,255:red,222;green,139;blue,175}, draw=none, bar shift=3.2pt, forget plot] coordinates {(Shank,89.52) [89.5]};
\node[font=\fontsize{4.5}{4.5}\selectfont, color=black!70, anchor=south, inner sep=0.8pt, xshift=3.2pt] at (axis cs:Shank,50) {R};
\addplot[fill={rgb,255:red,185;green,84;blue,127}, draw=none, bar shift=-3.2pt, forget plot] coordinates {(Foot,90.11) [90.1]};
\node[font=\fontsize{4.5}{4.5}\selectfont, color=white, anchor=south, inner sep=0.8pt, xshift=-3.2pt] at (axis cs:Foot,50) {L};
\addplot[fill={rgb,255:red,185;green,84;blue,127}, draw=none, bar shift=3.2pt, forget plot] coordinates {(Foot,89.93) [89.9]};
\node[font=\fontsize{4.5}{4.5}\selectfont, color=white, anchor=south, inner sep=0.8pt, xshift=3.2pt] at (axis cs:Foot,50) {R};
\addlegendimage{legend image code/.code={\fill[draw=none, fill={rgb,255:red,156;green,134;blue,210}] (0cm,-0.07cm) rectangle (0.2cm,0.11cm);}}
\addlegendentry{Trunk and head}
\addlegendimage{legend image code/.code={\fill[draw=none, fill={rgb,255:red,90;green,146;blue,209}] (0cm,-0.07cm) rectangle (0.2cm,0.11cm);}}
\addlegendentry{Arms}
\addlegendimage{legend image code/.code={\fill[draw=none, fill={rgb,255:red,222;green,139;blue,175}] (0cm,-0.07cm) rectangle (0.2cm,0.11cm);}}
\addlegendentry{Legs}
\addlegendimage{legend image code/.code={\draw[black!45, dashed, line width=0.5pt] (0cm,0.02cm) -- (0.36cm,0.02cm);}}
\addlegendentry{all sensors (93.6)}
\end{axis}
\end{tikzpicture}
\vspace{-6pt}
\caption{Placement on the 168 unseen sequences of Nymeria's test split: accuracy of the recognized body segment (17 classes) for sensors moved around their segment and remounted as in training (dashed: all sensors).}
\label{fig:place}
\end{figure}

\begin{table}[t]
\caption{Nymeria held-out protocol (3{,}908 windows). Top: IMU--text retrieval among 100 candidates and among all windows. Bottom: captioning. HiT: mean of three seeds of the language stage.}
\label{tab:nymeria_lang}
\centering
\small
\resizebox{\linewidth}{!}{%
\begin{tabular}{llcccc}
\toprule
& & \multicolumn{2}{c}{AnyMo} & \multicolumn{2}{c}{HiT (ours)} \\
\cmidrule(lr){3-4}\cmidrule(lr){5-6}
& & IMU$\to$T & T$\to$IMU & IMU$\to$T & T$\to$IMU \\
\midrule
\multirow{4}{*}{100} & R@1 & 28.0 & 33.0 & \textbf{42.7} & \textbf{41.7} \\
& R@5 & 63.0 & 60.0 & \textbf{78.0} & \textbf{76.3} \\
& R@10 & 77.0 & 79.0 & \textbf{86.7} & \textbf{85.7} \\
& MRR & 44.6 & 46.7 & \textbf{58.4} & \textbf{57.3} \\
\midrule
\multirow{2}{*}{all} & R@1 & 2.3 & 3.0 & \textbf{6.2} & \textbf{5.9} \\
& MRR & 7.0 & 7.5 & \textbf{13.8} & \textbf{13.2} \\
\bottomrule
\end{tabular}}

\vspace{4pt}
\setlength{\tabcolsep}{3pt}%
\resizebox{\linewidth}{!}{%
\begin{tabular}{lccccc}
\toprule
& BLEU-1 & BLEU-4 & ROUGE-L & METEOR & BERT-F1 \\
\midrule
AnyMo & 25.0 & 6.5 & 31.1 & 33.5 & 69.7 \\
HiT (ours) & \textbf{35.8} & \textbf{11.0} & \textbf{36.5} & \textbf{44.7} & \textbf{71.5} \\
\bottomrule
\end{tabular}}
\end{table}

\begin{table}[t]
\caption{Zero-shot Ego-Exo4D.
Retrieval among 100 candidates, IMU$\to$text / text$\to$IMU; ``MRR (cap.)'' retrieves with the bge-large embedding of the generated caption, so it also applies to IMU4D, which has no IMU embedding.
IMU4D is rescored by us on the same windows; the other baselines are the numbers reported by AnyMo; HiT is the mean of three seeds of the language stage.}
\label{tab:egoexo}
\centering
\footnotesize
\setlength{\tabcolsep}{2.5pt}
\begin{tabular}{@{}lcccccc@{}}
\toprule
& \multicolumn{4}{c}{Retrieval (IMU$\to$text / text$\to$IMU)} & \multicolumn{2}{c}{Captioning} \\
\cmidrule(lr){2-5}\cmidrule(l){6-7}
& R@1 & R@10 & MRR & MRR (cap.) & METEOR & BERT-F1 \\
\midrule
AnyMo & 2/3 & 27/23 & 9.5/9.9 & 7.7/7.3 & 30.3 & 67.1 \\
IMU4D & -- & -- & -- & 5.1/6.8 & 24.9 & \textbf{67.5} \\
HiT (ours) & \textbf{9/6} & \textbf{45/44} & \textbf{20.6/18.1} & \textbf{12.2/11.5} & \textbf{42.2} & 63.4 \\
\bottomrule
\end{tabular}
\end{table}
\begin{figure}[t]
\centering
\definecolor{gRef}{HTML}{C8C8C8}
\definecolor{gAnyMo}{HTML}{7F8FA9}
\definecolor{gIMUfourD}{HTML}{DB9E61}
\definecolor{gHiTlang}{HTML}{A3D4B2}
\definecolor{gHiT}{HTML}{4FA872}
\begin{tikzpicture}
\begin{axis}[
  xbar=0.9pt, bar width=5.4pt,
  width=0.80\linewidth, height=3.2cm, scale only axis,
  xmin=40, xmax=90, xtick={40,50,60,70,80,90},
  xlabel={Correct (\%)}, xlabel style={font=\scriptsize, yshift=3pt},
  xticklabel style={font=\scriptsize},
  symbolic y coords={Posture,Acting hand}, ytick={Posture,Acting hand},
  yticklabel style={font=\footnotesize, align=center}, yticklabels={Posture,Acting\\hand},
  enlarge y limits=0.36,
  axis x line*=bottom, axis y line*=left, tick align=outside, major tick length=2pt,
  xmajorgrids, grid style={black!10},
  point meta=explicit symbolic,
  nodes near coords, nodes near coords style={font=\fontsize{5.5}{5.5}\selectfont, color=black!85, anchor=west, inner sep=1.5pt},
  legend cell align=left, reverse legend,
  legend style={at={(0.5,1.03)}, anchor=south, legend columns=3, font=\scriptsize, draw=none, fill=none,
    /tikz/every even column/.append style={column sep=5pt}},
  legend image code/.code={\fill[#1] (0cm,-0.07cm) rectangle (0.2cm,0.11cm);},
]
\addplot[draw=none, fill=gHiT] coordinates {(81.0,Acting hand) [81.0] (68.5,Posture) [68.5]};
\addplot[draw=none, fill=gHiTlang] coordinates {(64.8,Acting hand) [64.8] (65.2,Posture) [65.2]};
\addplot[draw=none, fill=gIMUfourD] coordinates {(56.5,Acting hand) [56.5] (49.5,Posture) [49.5]};
\addplot[draw=none, fill=gAnyMo] coordinates {(45.2,Acting hand) [45.2$^*$] (58.5,Posture) [58.5]};
\addplot[draw=none, fill=gRef] coordinates {(82.9,Acting hand) [82.9] (83.0,Posture) [83.0]};
\legend{HiT, {HiT, language only}, IMU4D, AnyMo, Human reference}
\end{axis}
\end{tikzpicture}
\vspace{-6pt}
\caption{
Correct statements (\%) on the Nymeria held-out windows whose caption makes them, checked against ground-truth SMPL: the hand that moves more, posture (standing, walking, sitting, on the floor, lying).
Human reference: Nymeria's own narrations of the same windows. HiT and its language-only variant: means of three seeds.  $^*$AnyMo names a hand in 42 of 3{,}908 windows.}
\label{fig:ground}
\end{figure}
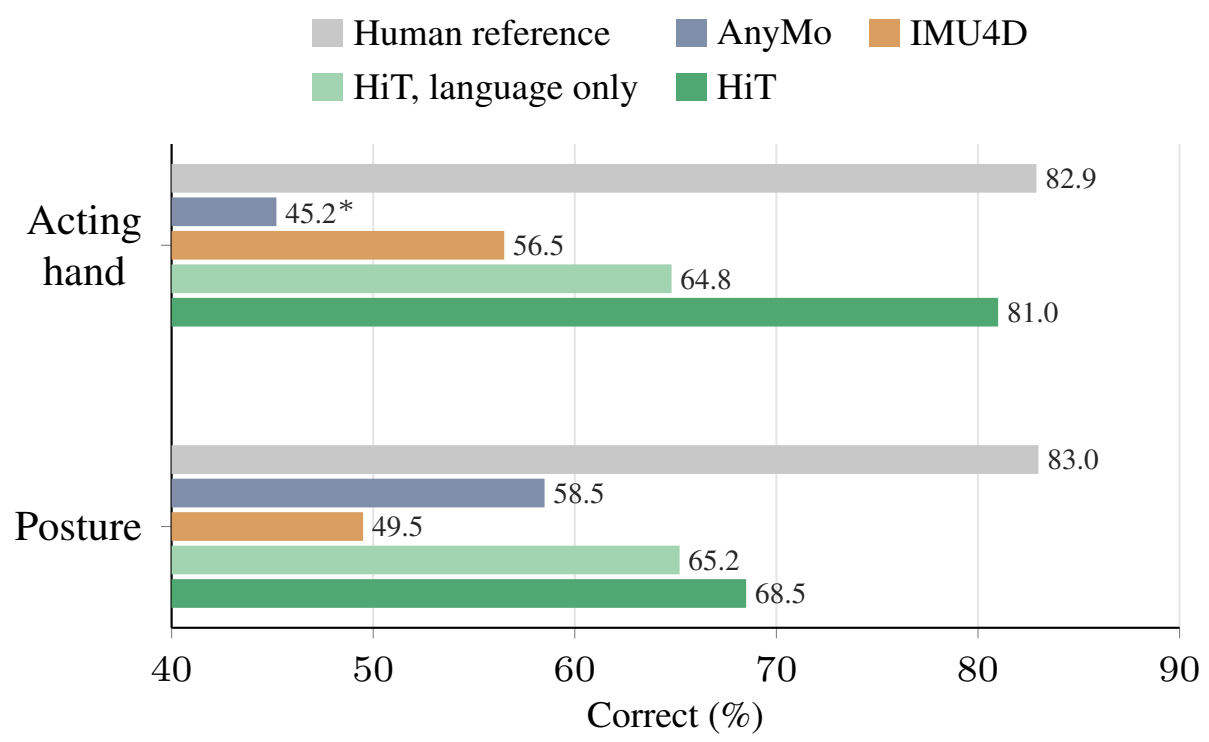

\paragraph{Results.}
HiT improves retrieval and captioning on the Nymeria held-out protocol (R@1 among 100 candidates 28/33 to 43/42, MRR over all windows 7.0/7.5 to 13.8/13.2, BLEU-4 6.5 to 11.0; \cref{tab:nymeria_lang}) and on zero-shot Ego-Exo4D~\cite{egoexo4d} (MRR 9.5/9.9 to 20.6/18.1; \cref{tab:egoexo}).
IMU4D's BERT-F1 is higher, but its captions are short and generic (13 words on average, about the length of the references, against 39 for HiT): retrieving the window from the generated caption works about twice as well with HiT's captions.
In \cref{fig:captions}, HiT names the action and the limb that performs it, while AnyMo's captions stay generic and IMU4D's describe other motions; activities absent from Nymeria are mapped to ones of training (playing badminton in 38\% of the basketball windows, handling a pan in 25\% of the bicycle-repair windows).

\subsection{Pose estimation}
\label{sec:exp:pose}

\paragraph{Results.}
HiT achieves competitive pose accuracy without sensor calibration or known sensor locations (\cref{tab:pose}).
In the zero-shot setting, it outperforms IMU4D on both IMUPoser and
DIP-IMU, reducing MPJPE from 15.03 to 11.24\,cm and from 8.78 to
7.82\,cm, respectively.
On TotalCapture, HiT achieves the lowest PA-MPJPE, although IMU4D
and ProbIP~\cite{probip} obtain lower MPJPE using calibrated
orientations; ProbIP additionally starts from the ground-truth pose.
Fine-tuning makes HiT the most accurate method on IMUPoser,
while IMU4D retains a small advantage on DIP-IMU and TotalCapture.
These comparisons also involve a difference in supervision:
HiT's pre-training uses only Nymeria, whose SMPL labels are
retargeted from a motion-capture suit, whereas the test sets and
IMU4D's pre-training data use AMASS-style SMPL fits (Supplementary~A).

HiT's clearest advantage is that its accuracy is largely preserved
when the sensors are not calibrated.
Under simulated remounting, with rotations of up to $180^\circ$
about the skin normal and tilts of up to $15^\circ$, its zero-shot
MPJPE on DIP-IMU changes only from 7.82 to 7.88\,cm, whereas
IMU4D's rises from 8.78 to 25.97\,cm.
This robustness persists after fine-tuning.
The calibration experiment on IMUPoser shows the same pattern:
across twelve device combinations, removing calibration raises
MobilePoser's MPJPE from 8.15 to 23.14\,cm, while HiT's accuracy
is essentially unchanged in both zero-shot and fine-tuned settings.
Together, these results show that HiT maintains its pose accuracy
under the mounting and calibration variations tested here
(\cref{fig:qual}).

The remaining trade-off is temporal smoothness.
HiT's raw predictions have higher jitter than MobilePoser's,
but applying the same zero-phase 3\,Hz filter used by IMU4D
reduces HiT's jitter on DIP-IMU from 1.04 to 0.21\,$\times10^2$\,m/s$^3$ without
changing MPJPE.
Beyond pose, HiT predicts root velocity and yaw rate to recover
the root trajectory.
Its zero-shot trajectory error is less than half of IMU4D's
on IMUPoser (0.17 vs.\ 0.41\,m) and the lowest on TotalCapture;
its advantage over IMU4D on IMUPoser also remains after
fine-tuning (Supplementary~B).

\subsection{Sensor placement}
\label{sec:exp:place}

HiT is not told where a sensor is worn; its placement head estimates it.
On the unseen sequences of Nymeria's test split it recognizes the segment of 93.6\% of the sensors (17 classes, \cref{fig:place}; 89.3\% from a single sensor), with a median error of $6.9^\circ$ for the angle around the segment and $8.6^\circ$ for the device orientation; the angle error is $4.4^\circ$ in the most certain quarter of the estimates and $30.4^\circ$ in the least certain.
Zero-shot on the pose test sets, it recognizes every sensor of TotalCapture, 91--100\% of DIP-IMU's sensors and 96--100\% of IMUPoser's watches, but a pocket phone as the thigh in only 54--72\% of windows (86--100\% counting the adjacent pelvis) and an earbud as the head in 38--76\%.
Fine-tuning with the segment loss, as for HiT's fine-tuned rows of \cref{tab:pose}, removes these errors (98--100\% on all three sets) at an unchanged pose error.

\subsection{Ablations}
\label{sec:exp:abl}
\label{sec:exp:posesup}

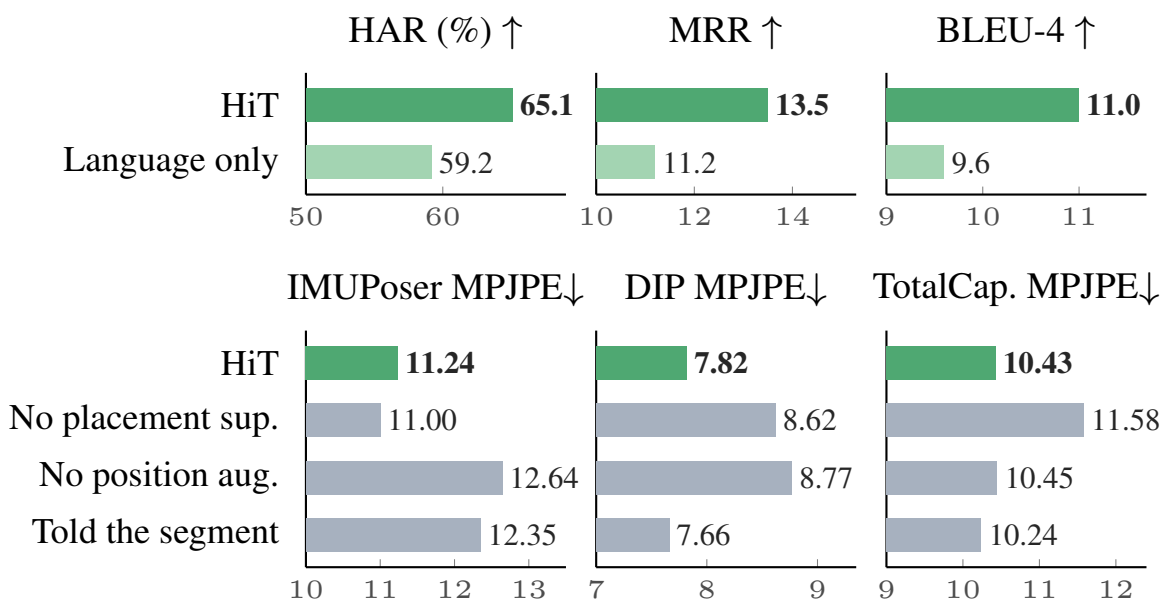
\begin{figure}[t]
\centering
\definecolor{aHiT}{HTML}{4FA872}
\definecolor{aLang}{HTML}{A3D4B2}
\definecolor{aAbl}{HTML}{A7B1BF}
\pgfplotsset{ablpanel/.style={
  xbar, bar shift=0pt, bar width=6.3pt, width=1.72cm, scale only axis,
  axis y line*=left, axis x line*=bottom,
  xticklabel style={font=\fontsize{5.3}{5.3}\selectfont, color=black!70, yshift=1pt}, xtick style={black!50}, major tick length=1.5pt,
  /pgf/number format/fixed,
  yticklabel style={font=\fontsize{6.8}{6.8}\selectfont}, ytick style={draw=none},
  title style={font=\fontsize{7}{8}\selectfont, yshift=-5pt},
  point meta=explicit symbolic,
  nodes near coords, nodes near coords style={font=\fontsize{5.8}{5.8}\selectfont, color=black!85, anchor=west, inner sep=1.2pt},
}}
\begin{tikzpicture}
\begin{groupplot}[group style={group size=3 by 2, horizontal sep=0.2cm, vertical sep=0.9cm}, ablpanel]
\nextgroupplot[title={HAR (\%) $\uparrow$}, xmin=50, xmax=69, xtick={50,60}, ymin=0.44, ymax=2.56, height=0.805cm, ytick={2,1}, yticklabels={HiT,Language only}]
\addplot[fill=aHiT, draw=none] coordinates {(65.1,2) [\textbf{65.1}]};
\addplot[fill=aLang, draw=none] coordinates {(59.2,1) [59.2]};
\nextgroupplot[title={MRR $\uparrow$}, xmin=10, xmax=15.3, xtick={10,12,14}, ymin=0.44, ymax=2.56, height=0.805cm, ytick={2,1}, yticklabels={}]
\addplot[fill=aHiT, draw=none] coordinates {(13.5,2) [\textbf{13.5}]};
\addplot[fill=aLang, draw=none] coordinates {(11.2,1) [11.2]};
\nextgroupplot[title={BLEU-4 $\uparrow$}, xmin=9, xmax=11.7, xtick={9,10,11}, ymin=0.44, ymax=2.56, height=0.805cm, ytick={2,1}, yticklabels={}]
\addplot[fill=aHiT, draw=none] coordinates {(11.0,2) [\textbf{11.0}]};
\addplot[fill=aLang, draw=none] coordinates {(9.6,1) [9.6]};
\nextgroupplot[title={IMUPoser MPJPE$\downarrow$}, xmin=10, xmax=13.5, xtick={10,11,12,13}, ymin=0.44, ymax=4.5600000000000005, height=1.565cm, ytick={4,3,2,1}, yticklabels={HiT,No placement sup.,No position aug.,Told the segment}]
\addplot[fill=aHiT, draw=none] coordinates {(11.24,4) [\textbf{11.24}]};
\addplot[fill=aAbl, draw=none] coordinates {(11.0,3) [11.00]};
\addplot[fill=aAbl, draw=none] coordinates {(12.64,2) [12.64]};
\addplot[fill=aAbl, draw=none] coordinates {(12.35,1) [12.35]};
\nextgroupplot[title={DIP MPJPE$\downarrow$}, xmin=7, xmax=9.35, xtick={7,8,9}, ymin=0.44, ymax=4.5600000000000005, height=1.565cm, ytick={4,3,2,1}, yticklabels={}]
\addplot[fill=aHiT, draw=none] coordinates {(7.82,4) [\textbf{7.82}]};
\addplot[fill=aAbl, draw=none] coordinates {(8.62,3) [8.62]};
\addplot[fill=aAbl, draw=none] coordinates {(8.77,2) [8.77]};
\addplot[fill=aAbl, draw=none] coordinates {(7.66,1) [7.66]};
\nextgroupplot[title={TotalCap.\ MPJPE$\downarrow$}, xmin=9, xmax=12.4, xtick={9,10,11,12}, ymin=0.44, ymax=4.5600000000000005, height=1.565cm, ytick={4,3,2,1}, yticklabels={}]
\addplot[fill=aHiT, draw=none] coordinates {(10.43,4) [\textbf{10.43}]};
\addplot[fill=aAbl, draw=none] coordinates {(11.58,3) [11.58]};
\addplot[fill=aAbl, draw=none] coordinates {(10.45,2) [10.45]};
\addplot[fill=aAbl, draw=none] coordinates {(10.24,1) [10.24]};
\end{groupplot}
\end{tikzpicture}
\vspace{-4pt}
\caption{Ablations of the backbone.
Top: zero-shot HAR accuracy (\%, ten datasets), retrieval MRR (all Nymeria held-out windows, both directions) and Nymeria BLEU-4 of HiT and of the same model trained on language alone (means of three seeds).
Bottom: zero-shot MPJPE (cm, \cref{tab:pose}) of the input and supervision ablations (HAR accuracy: Supplementary Tab.~7).}
\label{fig:abl}
\end{figure}

All ablations use HiT's segment output and caption decoder; HiT's three seeds range over 63.6--66.2\% HAR accuracy and 12.5--14.3 MRR.

\paragraph{Pose supervision.}
Training the backbone for pose and placement before freezing it
improves all three language outputs over the same model trained on language alone (\cref{fig:abl}).
Zero-shot HAR accuracy rises from 59.2\% to 65.1\%, a gain more than twice
the range across HiT's seeds.
In all nine pairs of seeds, the language-only model is
significantly worse (McNemar's test, $p<0.05$) on five to seven of ten datasets and significantly better only on USC-HAD and, in two pairs, UCI-HAR.
Retrieval MRR rises from 11.2 to 13.5 and BLEU-4 from 9.6 to 11.0.
Most of the gain comes from pose supervision itself: without the placement losses, the model still reaches 63.8\% (Supplementary Tab.~7).

\paragraph{Caption grounding.}
Caption metrics do not tell whether a caption describes the actual motion, so we check statements about the acting hand and posture against ground-truth SMPL on the held-out windows, using Nymeria's own narrations as a human reference. Acting-hand accuracy distinguishes the two training regimes (\cref{fig:ground}): among captions that name an acting hand, HiT identifies it correctly in 81.0\% of cases (70.8--87.3\% across seeds), comparable to the human reference and higher than the language-only model's 64.8\% (63.7--66.1\%). HiT and its language-only variant also outperform AnyMo and IMU4D in posture accuracy.

\paragraph{Placement supervision, placement-free input and position augmentation.}
All three mainly serve pose (\cref{fig:abl}).
Placement supervision reduces the pose error on DIP-IMU and TotalCapture, though not on IMUPoser.
Explicit segment labels help when they match the test setup, but can hurt when they misrepresent how a device is worn or moves:
our ``told'' control, which replaces the shared stem embedding with a learned body-segment embedding (true segments in training, benchmark descriptions at test time; Supplementary~C), slightly improves pose on DIP-IMU and TotalCapture, whose sensors are named by their site, but raises the IMUPoser MPJPE from 11.24 to 12.35\,cm.
Without position augmentation, \ie with every sensor where the suit has it (mounting augmentation kept), the pose error rises on IMUPoser (12.64 vs.\ 11.24\,cm) and DIP-IMU (8.77 vs.\ 7.82\,cm), not on TotalCapture.
Placement-free input thus avoids depending on how device descriptions map to body segments, and position augmentation prepares the model for devices that sit elsewhere around a limb than the suit's sensors.

\section{Conclusion}
\label{sec:conclusion}

HiT estimates full-body pose from any subset of body-worn IMUs, without being told where they are worn, and describes the motion in language.
Trained with random mounting rotations, HiT needs no calibration: its pose is unchanged when every device is remounted with a different spin about the skin normal (up to $180^\circ$, with up to $15^\circ$ of tilt), whereas the error of calibrated methods doubles or triples.
Its placement head recognizes the body segment of 93.6\% of the sensors on unseen sequences and estimates how each device sits on it.
Outputting language from the frozen model trained for pose and placement, instead of training the backbone on language end-to-end, raises zero-shot activity recognition from 59.2\% to 65.1\%, retrieval MRR from 11.2 to 13.5 and BLEU-4 from 9.6 to 11.0 with the same narrations; with the time-segment output and its three-way supervision, HiT improves zero-shot recognition on ten unseen datasets over the previous best of 43.9\% and retrieval and captioning over the state of the art.

\paragraph{Limitations and Future Work.}
The invariance covers rotations about the skin normal and small tilts, not arbitrary device orientations, and sensors with no axis along the skin normal cost accuracy; with glasses and wristbands alone, the legs are unobserved.
Consumer devices outside the training data are placed less reliably without fine-tuning (an earbud is often read as a hand).
HiT is an offline model: it attends over a whole window of up to 30\,s and is not causal; a real-time variant is left to future work.

\section*{LLM Usage Statement}

Large language models (Claude, Anthropic) were used to edit the prose of this paper and to assist with code for data processing and plotting. The experiments, their design, the results and the claims are the authors' own, and every generated passage was checked by the authors.

{
    \small
    \bibliographystyle{ieeenat_fullname}
    \bibliography{main}
}

\clearpage
\maketitlesupplementary
\appendix
\section{Pose evaluation protocol}
\label{sec:supp:pose}

This section specifies the evaluation protocol and per-method settings used to produce Tab.~1 of the main paper.

\subsection*{Data, splits, and metrics}
\paragraph{Test sets.}
IMUPoser~\cite{imuposer} follows the MobilePoser~\cite{mobileposer} protocol and evaluator: test participants P9--P10, its 12 device combinations (2, 6 and 4 combinations with 1, 2 and 3 devices, each with a phone in a front pocket), the original SMPL ground truth, and each pair of recording and combination counted once. DIP-IMU~\cite{dip} uses the IMU4D~\cite{imu4d} samples of the test subjects s\_09 and s\_10 (19 recordings, 30\,Hz) with five IMUs (both thighs, head and both wrists), and IMU4D's refitted SMPL-X ground truth. TotalCapture~\cite{totalcapture} uses the SMPL ground truth of the TransPose~\cite{transpose} release, at 30\,Hz, with three IMUs (head and both wrists).

\paragraph{Metrics.}
On DIP-IMU and TotalCapture we report the TransPose metrics, frame-weighted over sequences: MPJPE (cm), MPJRE (degrees), and PA-MPJPE (cm; per-frame similarity alignment of the 24 joints).
In each group, the best value of a column is in bold and the second best underlined among the methods that use the column's sensors.

\subsection*{Inputs and calibration}
\paragraph{Calibration and remounting.}
HiT reads device-frame specific force and angular velocity (one axis is taken as the skin normal) and is not given sensor-site labels. MobilePoser's released model, ProbIP~\cite{probip} and IMU4D instead consume calibrated inputs: device orientations aligned to the bones by a T-pose and to a common world frame.
To remove the calibration of IMUPoser we apply an unknown world heading and an unknown mounting rotation to every device (up to $180^\circ$ about the skin normal and up to $15^\circ$ of tilt), identically for all methods. This raises the MPJPE of MobilePoser's released model from 8.15 to 23.14\,cm, while that of HiT changes from 11.24 to 11.29\,cm (zero-shot) and from 6.30 to 6.32\,cm (fine-tuned; 6.29 in Tab.~1 of the main paper, from a separate evaluation run of the same model).

\paragraph{Device signals of the pose test sets.}
The releases used here for IMUPoser, DIP-IMU and TotalCapture provide calibrated orientations and gravity-free world-frame accelerations, but no gyroscope signals. We reconstruct device-frame specific force by restoring gravity and rotating into the recorded frame, and angular velocity by finite-differencing orientations at the native 25--30\,Hz before resampling to 60\,Hz. Both are invariant to the choice of world heading.

These reconstructed frames are calibrated and bone-aligned, so Tab.~1 of the main paper alone does not establish calibration independence. Our simulated remounting tests apply unknown mounting rotations to the recorded orientations before reconstruction; MPJPE remains at 11.29\,cm on IMUPoser and 7.88\,cm on DIP-IMU (Sec.~4.4 of the main paper). For these releases, we select a fixed skin-normal axis by wear location because the original device frames are unavailable. In deployment, this axis follows the device geometry (e.g., the normal of a watch face or phone screen). Neither wear location nor device type is supplied to the model.

The Nymeria~\cite{nymeria} glasses and wristbands, the ten activity benchmarks and Ego-Exo4D~\cite{egoexo4d} provide raw device-frame accelerometer and gyroscope signals, which HiT uses directly. Nymeria's suit provides accelerometer readings, with angular velocity derived from fused orientations (Supplementary~\ref{sec:supp:train}). Real-device removal and remounting with synchronized raw IMU and pose ground truth remain untested; our simulated remounting and tilt experiments serve as controlled proxies.

\subsection*{Per-method settings for Tab.~1 of the main paper}
\paragraph{HiT (ours).}
We use the same Nymeria-trained stage~1 model as in the other evaluations (Sec.~4.1 of the main paper).
Stage~2 freezes the pose estimation modules, so language training does not change the pose predictions.
Sensor-site labels are not provided as inputs.

\paragraph{Fine-tuning HiT.}
We fine-tune all backbone and head parameters for 1{,}875 steps with AdamW (learning rate $5\times10^{-5}$, no weight decay, 100 warm-up steps, gradient clipping at 1) without augmentation.
Each step contains 320\,s of signal, with a window length sampled from 5, 10, 20, and 30\,s and a dataset-specific sensor set sampled per window. We evaluate the final checkpoint without validation-based selection. The objective combines the pose loss with segment-classification cross-entropy (weight 0.5), assigning watches and wrist sensors to the forearm, pocket phones to the thigh, and earbuds to the head. We supervise yaw rate on all datasets and root velocity where translation is available. Translation comes from IMUPoser's annotations and TotalCapture's raw Vicon hip positions, shifted by one frame and rotated $180^\circ$ about the vertical to match the release.
We omit velocity supervision for two TotalCapture recordings with inconsistent turning and for DIP-IMU, which lacks translation. Training uses IMUPoser P1--P8 with all 24 device sets and original SMPL annotations; IMU4D's DIP-IMU split (s\_01--s\_08, 41 recordings) with refitted annotations; and TotalCapture S1--S3 excluding test motions (27 recordings), with the released SMPL annotations.

\paragraph{MobilePoser.}
The released model, the one shown in Fig.~4 of the main paper, is pre-trained on AMASS~\cite{amass} and fine-tuned on DIP-IMU and IMUPoser P1--P8. It is not zero-shot on IMUPoser or DIP-IMU, so its zero-shot row has only the TotalCapture cells.
On DIP-IMU it reads the calibrated orientations and accelerations of the five IMUs, as in its own evaluation; it is trained on DIP-IMU's original SMPL ground truth, so scoring against IMU4D's refitted ground truth disadvantages it.
On TotalCapture we fine-tune its joints and poser networks on S1--S3 with HiT's recipe and its own losses, with the head and wrists. Its zero-shot TotalCapture results use the head and both wrists. Each of its twelve training combinations includes a pocket phone. Using its three-device combinations with lower-leg sensors in place of the pocket phone gives zero-shot errors of 11.06\,cm MPJPE, $23.8^\circ$ MPJRE and 8.83\,cm PA-MPJPE on TotalCapture. Its AMASS pre-training contains TotalCapture's motions as synthetic IMU data.

\paragraph{ProbIP.}
We use the released three-sensor model (wrists and head), which requires calibrated orientations and the ground-truth initial pose. Its sensor sets differ from our evaluation protocol on IMUPoser and DIP-IMU, so these results are grey and unranked.
Since its training includes DIP-IMU, we report results there only in the trained group. Evaluation against IMU4D's refitted ground truth also disadvantages the model, which was trained on the original SMPL annotations. On IMUPoser and TotalCapture, we fine-tune it with its own pose losses and the same training budget and learning rate as HiT.

\paragraph{IMU4D.}
We use the released pre-trained model for zero-shot evaluation and dataset-specific fine-tuned models for IMUPoser and DIP-IMU.
On TotalCapture, we fine-tune on S1--S3 using IMU4D's real-world recipe. We retain its heading alignment and 3\,Hz low-pass filter, including ground-truth pelvis yaw on IMUPoser.
Its pre-training data contain motions from TotalCapture test recordings, suggesting prior exposure through synthetic IMU data. The IMUPoser checkpoint was fine-tuned on IMU4D's SMPL-X refit, whereas we evaluate against the original SMPL annotations, disadvantaging this checkpoint.
However, HiT's zero-shot advantage persists under either ground truth: using IMU4D's test samples, five devices, and 22-joint MPJPE, IMU4D obtains 14.1\,cm on both, versus HiT's 7.8\,cm on the refit and 7.5\,cm on the original annotations.

\subsection{Additional pose results}
\label{sec:supp:pose:additional}

\paragraph{Tilt beyond the training range.}
The mounting augmentation tilts a device by at most $15^\circ$ (Sec.~3.2 of the main paper). We evaluate robustness to larger tilts by applying, to every sensor of a sequence, a random spin (up to $180^\circ$) about the skin normal and then a fixed tilt of $t$ about a random axis in the skin plane (one draw per sequence), zero-shot with the inputs of Tab.~1 of the main paper.
At $90^\circ$, MPJPE increases by 1.2\,cm on DIP-IMU and 0.7\,cm on TotalCapture compared with $t=0$.

\begin{table}[t]
\caption{Pose on IMUPoser against IMU4D by number of devices, under MobilePoser's protocol and evaluator (test participants P9--P10; its 12 device combinations, each with a phone in a front pocket: 2, 6 and 4 combinations with 1, 2 and 3 devices; MPJPE in cm, MPJRE in degrees; PA-MPJPE in cm, after a per-frame similarity alignment of the 24 joints).
IMU4D reads its calibrated device orientations and is told each device's site; HiT uses reconstructed device-frame specific force and angular velocity without sensor-site labels.}
\label{tab:pose_imuposer}
\centering
\small
\resizebox{\linewidth}{!}{%
\begin{tabular}{l|cc|cc|cc}
\toprule
& \multicolumn{2}{c|}{MPJPE$\downarrow$} & \multicolumn{2}{c|}{MPJRE$\downarrow$} & \multicolumn{2}{c}{PA-MPJPE$\downarrow$} \\
Devices & IMU4D & HiT & IMU4D & HiT & IMU4D & HiT \\
\midrule
\multicolumn{7}{l}{\emph{Zero-shot}} \\
1 & 11.85 & \textbf{11.80} & \textbf{27.2} & 29.9 & 11.33 & \textbf{10.25} \\
2 & 14.68 & \textbf{11.44} & 29.4 & \textbf{28.6} & 12.38 & \textbf{8.91} \\
3 & 17.14 & \textbf{10.66} & 31.3 & \textbf{27.4} & 13.51 & \textbf{8.31} \\
\midrule
\multicolumn{7}{l}{\emph{Fine-tuned on IMUPoser P1--P8}} \\
1 & 8.96 & \textbf{7.77} & 24.0 & \textbf{19.8} & 8.59 & \textbf{7.32} \\
2 & 7.58 & \textbf{6.28} & 21.5 & \textbf{16.3} & 6.62 & \textbf{5.47} \\
3 & 7.12 & \textbf{5.58} & 20.8 & \textbf{14.6} & 5.93 & \textbf{4.57} \\
\bottomrule
\end{tabular}}
\end{table}

\paragraph{Other pose results.}
\Cref{tab:pose_imuposer} breaks the IMUPoser column of Tab.~1 of the main paper down by number of devices: zero-shot HiT stays at 10.7--11.8\,cm with one to three devices while IMU4D's error grows from 11.85 to 17.14\,cm.
On the Nymeria validation split the joint-angle error is $9.81^\circ$ with glasses and both wristbands, $12.33^\circ$ with the right wristband alone, and $7.63^\circ$ with six sensors (adding the suit's pelvis and thigh sensors).
On WHIP~\cite{whip} (watches and pocket phones), HiT's mean MPJPE over sensor sets is 14.62\,cm for the unseen actor and 15.01\,cm for the unseen actions; a constant pose gives 19.29 and 20.05\,cm.
On IMUPoser with both watches and the earbuds, removing calibration increases ProbIP's MPJPE from 9.11 to 23.55\,cm.

\paragraph{Training on AMASS.}
We also trained an earlier, pose-only variant on synthetic IMU signals from AMASS. This variant receives sensor-site labels; neither DIP-IMU nor TotalCapture is included in its training data. Averaged over two seeds, it achieves 7.68 / 23.0 / 6.09 on DIP-IMU and 7.75 / 17.5 / 5.92 on TotalCapture (zero-shot). On TotalCapture's 14 test recordings of the standard split, it obtains 8.6--8.7\,cm MPJPE, compared with 9.80\,cm for IMU4D.
With the same training setup and budget, the Nymeria-trained variant obtains 8.61\,cm MPJPE on DIP-IMU, compared with 7.68\,cm for the AMASS-trained variant.

\section{Root trajectory}
\label{sec:supp:traj}

\begin{table}[h]
\caption{Root trajectory error (m) on IMU4D's evaluation windows: ATE / FDE over all windows and over the windows whose ground-truth root moves at least 1\,m.
IMUPoser: P9--P10, five devices, 113 windows (43 moving); TotalCapture: head and both wrists, 456 windows (282 moving).}
\label{tab:traj}
\centering
\small
\resizebox{\linewidth}{!}{%
\begin{tabular}{lcccc}
\toprule
& \multicolumn{2}{c}{IMUPoser} & \multicolumn{2}{c}{TotalCapture} \\
\cmidrule(lr){2-3}\cmidrule(lr){4-5}
Method & All & Moving & All & Moving \\
\midrule
Not moving & 0.68 / 1.03 & 1.72 / 2.59 & 0.89 / 1.21 & 1.39 / 1.91 \\
\midrule
\multicolumn{5}{l}{\emph{Not trained on the target dataset}} \\
IMU4D & 0.41 / 0.67 & 0.93 / 1.55 & 0.33 / 0.57 & 0.42 / 0.75 \\
IMU4D, noise-trained & 0.32 / 0.51 & 0.78 / 1.25 & 0.24 / 0.40 & 0.36 / 0.60 \\
HiT (ours) & \textbf{0.17} / \textbf{0.28} & \textbf{0.39} / \textbf{0.64} & \textbf{0.18} / \textbf{0.33} & \textbf{0.27} / \textbf{0.50} \\
\midrule
\multicolumn{5}{l}{\emph{Trained on the target dataset}} \\
IMU4D & 0.10 / 0.17 & 0.21 / 0.37 & -- & -- \\
HiT (ours) & \textbf{0.08} / \textbf{0.13} & \textbf{0.17} / \textbf{0.27} & \textbf{0.14} / \textbf{0.25} & \textbf{0.21} / \textbf{0.38} \\
\bottomrule
\end{tabular}}
\end{table}

HiT's pose head estimates, at every frame, the root velocity in the body's heading frame and the yaw rate; integrating them gives the horizontal root trajectory relative to the start of a window, not in a world frame.
IMU4D outputs the root translation.
We compare both on IMU4D's own evaluation windows of 200 frames (IMUPoser at its native rate of about 25\,Hz, TotalCapture at 30\,Hz) with the inputs of Tab.~1 of the main paper.
The ground truth is IMU4D's window translation on IMUPoser and the Vicon hip position on TotalCapture; DIP-IMU has no translation.
We translate each trajectory to start at the origin and align it to the ground truth using a single yaw rotation fitted separately for each trajectory.
We apply the same alignment procedure to all methods.
ATE is the mean horizontal distance to the ground truth over the window, and FDE is the distance at its end.
Zero-shot, HiT's error is less than half of IMU4D's on IMUPoser and the lowest on TotalCapture, whose motions IMU4D has likely seen in pre-training (\cref{tab:traj}).
After fine-tuning with root-velocity supervision (Supplementary~\ref{sec:supp:pose}), HiT is more accurate than IMU4D's fine-tuned model on IMUPoser and improves on TotalCapture from 0.18 to 0.14\,m (we did not evaluate the trajectory of our TotalCapture fine-tune of IMU4D).

\section{Ablation setup}
\label{sec:supp:abl}
Unless stated otherwise, each ablation in Fig.~8 of the main paper changes one component while keeping the data, augmentation, architecture, and training settings fixed.
We use the main-paper protocols for zero-shot pose estimation, zero-shot HAR, retrieval, and captioning.
Stage~1 uses one seed per variant to evaluate pose. Stage~2 uses three seeds, and we average their language results.

\paragraph{No placement supervision.}
Stage~1 with $\lambda_{\text{p}} = 0$ in Eq.~(1) of the main paper.
The sensor queries, the placement head and the routing of the raw-signal skip stay in the network, but no segment, angle or orientation loss is applied, so the sensor descriptors, which route the skip and enter the caption prompt, are learned by the pose loss alone.

\paragraph{No position augmentation.}
Every sensor stays where the suit or the Aria device has it, in both stages; the mounting augmentation (spin about the skin normal, tilt and flip) and the placement losses are kept.

\paragraph{Language only.}
We randomly initialize the backbone, decoder, and placement-head pooling, and train them jointly with the stage~2 language heads using only the three language losses in Eq.~(2).
We use the same narrated windows, training split, and augmentation as stage~2, with no pose or placement supervision and no pose output.
Training runs for 20k steps at a learning rate of $3\times10^{-4}$ ($10^{-4}$ for LoRA), with checkpoint selection on the Nymeria validation split.
Across three seeds, the model averages 59.2\% HAR accuracy, 11.2 MRR, and 9.6 BLEU-4.
Training takes 4.5\,h on four A100s, compared with 9\,h for HiT's stage~1 and 1.6\,h for stage~2.
Keeping the decoder and placement head frozen at HiT's stage~1 weights gives similar results in one seed: 57.4\% HAR accuracy, 11.6 MRR, and 9.7 BLEU-4.

\paragraph{Pose pretraining on the same sequences.}
HiT uses 813 sequences (212.7\,h) for pose pretraining, while language training uses only the 569 narrated sequences.
To test whether the extra sequences explain HiT's gains, we restrict pose pretraining to these 569 sequences (147.0\,h of recordings, of which 141.9\,h are narrated), keeping all training settings unchanged.
Across three seeds, this model achieves 64.8\% HAR accuracy, 14.5 MRR, and 11.0 BLEU-4, compared with 65.1\%, 13.5, and 11.0 for HiT and 59.2\%, 11.2, and 9.6 for the language-only model.
Zero-shot pose errors also remain similar: 11.34, 8.08, and 10.39\,cm on IMUPoser, DIP-IMU, and TotalCapture, versus HiT's 11.24, 7.82, and 10.43\,cm.
Thus, pose pretraining remains beneficial without the additional sequences.
Per-dataset HAR results are in \cref{tab:abl_har}.

\begin{table*}[t]
\caption{Zero-shot activity recognition (\%) per dataset of the ablations of Fig.~8 of the main paper and of the control with stage~1 on the narrated sequences only: means over three seeds except where a single seed is indicated. The language-only and same-sequence pose pretraining controls use the same 569 source recordings.}
\label{tab:abl_har}
\centering
\small
\resizebox{\linewidth}{!}{%
\setlength{\tabcolsep}{4pt}
\begin{tabular}{lcccccccccc|c}
\toprule
& Opp. & UCI & w-HAR & RW & TNDA & PAMAP2 & USC & WISDM & DSADS & UTD & Mean \\
\midrule
HiT - three seeds & 73.6 & 66.6 & 76.0 & 81.8 & 99.8 & 74.0 & 54.6 & 48.1 & 51.7 & 24.5 & 65.1 \\
\midrule
HiT - seed 0 & 74.1 & 66.1 & 80.3 & 84.6 & 99.9 & 75.4 & 60.8 & 48.5 & 48.6 & 23.3 & 66.2 \\
Language-only training & 60.5 & 66.3 & 64.5 & 66.5 & 88.1 & 68.0 & 59.2 & 47.1 & 46.4 & 25.0 & 59.2 \\
Pose pretraining on the same sequences & 72.6 & 69.4 & 72.7 & 77.0 & 99.9 & 73.6 & 58.2 & 50.1 & 48.4 & 26.4 & 64.8 \\
w/o placement supervision & 68.3 & 68.0 & 79.8 & 80.2 & 96.2 & 72.6 & 52.4 & 47.6 & 47.9 & 24.8 & 63.8 \\
w/o position augmentation & 73.5 & 59.7 & 71.6 & 82.0 & 100.0 & 73.7 & 51.1 & 52.2 & 49.2 & 22.0 & 63.5 \\
Sensor body-segment labels as input & 71.9 & 76.9 & 79.8 & 83.5 & 100.0 & 75.5 & 36.5 & 47.3 & 43.7 & 27.1 & 64.2 \\

\bottomrule
\end{tabular}}
\end{table*}

\paragraph{Placement as input.}
We replace HiT's shared sensor embedding with a learned segment embedding (17 classes), while retaining the placement head and its losses.
Training uses each sensor's known segment, which remains unchanged by position augmentation.
At test time, we map each benchmark's device locations to segments as follows: watches and wristbands to the forearm; ankle and shin sensors to the lower leg; waist and lower-back sensors to the pelvis; pocket phones to the upper leg; and chest, foot, and head devices to their corresponding segments.
Only segment identity is provided, without mounting angle or orientation.

\paragraph{Per-dataset recognition.}
\Cref{tab:abl_har} gives the accuracy of each ablation per dataset.
Gravity directions during labeled postures suggest two mismatches in the assigned segment labels.
USC-HAD's hip-pocket sensor remains upright during sitting, consistent with the pelvis, while DSADS's leg sensors become horizontal, consistent with the thighs.
The placement-as-input variant shows its largest accuracy drops on these two datasets.

\section{Supervision of the output}
\label{sec:supp:loss}

\begin{table*}[t]
\caption{Supervision of the output: narration alignment, activity labels and captioning, alone and combined, on the same frozen pose backbone.
Rows with captions use HiT's caption decoder; the last row is HiT with its first seed (its three seeds range over 63.6--66.2\% HAR accuracy, 36--46 / 39--44 R@1 and 10.5--11.5 BLEU-4); one seed per row.
HAR F1 as in Tab.~2 of the main paper; MRR over all held-out windows and R@1 among 100 candidates, IMU$\to$text / text$\to$IMU.
``--'': not available, because the output head has no captioning head or no alignment embedding. Results are shown with seed 0.}
\label{tab:loss}
\centering
\small
\begin{tabular}{ccc|cccccc}
\toprule
Align & Labels & Captions & HAR acc & HAR F1 & MRR & R@1 & BLEU-4 & BERT-F1 \\
\midrule
\checkmark & & & 64.4 & 57.4 & 14.5 / \textbf{14.2} & 40 / \textbf{45} & -- & -- \\
& \checkmark & & 57.3 & 54.0 & 4.7 / 4.6 & 15 / 18 & -- & -- \\
& & \checkmark & -- & -- & -- & -- & 10.7 & 71.4 \\
\checkmark & \checkmark & & 64.1 & 58.7 & 13.9 / 13.7 & 48 / 44 & -- & -- \\
\checkmark & & \checkmark & 65.5 & 58.0 & 13.6 / 13.9 & \textbf{49} / 43 & \textbf{11.4} & \textbf{72.1} \\
& \checkmark & \checkmark & 56.4 & 52.6 & 4.1 / 4.6 & 19 / 14 & 10.5 & 71.9 \\
\midrule
\checkmark & \checkmark & \checkmark & \textbf{66.2} & \textbf{59.4} & \textbf{14.6} / 14.0 & 36 / 44 & 11.1 & 71.9 \\
\bottomrule
\end{tabular}
\end{table*}

\Cref{tab:loss} compares narration alignment, activity classification, and captioning losses on the same frozen pose backbone.
Models trained with alignment achieve 64.1--66.2\% HAR accuracy and MRR of 13.6--14.6. The two variants without alignment that report these metrics achieve 56.4--57.3\% accuracy and MRR below 5.
Captioning alone reaches 10.7 BLEU-4, compared with 11.1 with all three losses.
Adding labels and captioning to alignment changes HAR accuracy by less than the variation across HiT's three seeds.

\section{Zero-shot activity recognition protocol}
\label{sec:supp:har}
Each benchmark contributes only its test windows (at most 10\,s), its sensor placements and its class names.
Signals are fed in the devices' own frames (device $z$ axis as the skin normal, 60\,Hz), and the model is not told where any sensor is worn.
The class names are inserted into three templates (``C is \{\}.'', ``C does this activity: \{\}.'', ``C is doing \{\} in the room.''); their projected bge-large~\cite{bge} embeddings are averaged over the templates, and the class most similar to the window's alignment vector is predicted.
Macro-F1 averages over the classes present in a dataset's test labels or predictions, and R@2 counts the true class among the two highest-scoring ones, as in the AnyMo~\cite{anymo} evaluation code.
Full-shot fine-tuning (Tab.~2 of the main paper, bottom) uses the UniMTS~\cite{unimts} training splits and recipe and fine-tunes all parameters of HiT for 20 epochs.
We report AnyMo's published results in Tab.~2 of the main paper. Running the released model with its own data loader and all of its sensors reproduces the mean accuracy of 43.9\% (macro-F1 35.4\% and R@2 68.0\% against the published 36.6\% and 69.7\%, a difference that comes from the 61 test windows of w-HAR).

\section{Architecture details}
\label{sec:supp:arch}
All attention blocks are pre-norm, with width $d = 512$, 8 heads and an MLP of width $4d$.

\paragraph{Backbone.}
A two-layer MLP embeds each patch of 15 samples into $d = 512$.
The backbone has 12 pairs of local and global blocks, with outputs taken after blocks 6, 12, 18, and 24.
All sensors share a single learned embedding.
Each patch holds one token per present sensor, a learned frame token and two learned registers.
A local block attends among the tokens of one patch; a global block attends over all tokens of the window, with rotary embeddings of the patch time.

\paragraph{Decoder.}
For patch $p$, the queries are five learned pose queries (torso, left arm, right arm, left leg, right leg) and one query per present sensor, $\mathbf{q}_{i,p} = \mathbf{q}_{\text{s}} + W \operatorname{LN}(\mathbf{h}_{i,p})$, where $\mathbf{q}_{\text{s}}$ is a learned vector shared by all sensors and $\mathbf{h}_{i,p}$ is the token of sensor $i$ after the last block.
The keys are the frame token and the sensor tokens of patches $p-1$, $p$ and $p+1$ at the four tapped depths, normalized per depth and given learned embeddings of their depth, of their kind (frame or sensor token) and of their patch offset, plus one learned null key.
Each layer applies self-attention among the $5 + S$ queries, cross-attention to the keys (absent sensors and padded patches are masked in both) and an MLP.
The pose queries go to the pose head and the sensor queries to the placement head.

\paragraph{Placement head.}
A linear score of each sensor query, normalized by a softmax over the patches where the sensor is present, pools the sensor's queries over the window; a linear layer and a GELU give its descriptor $\mathbf{d}_i$.
From $\mathbf{d}_i$, linear layers give the 17 segment logits and the angle around the segment as a von Mises distribution (a unit mean direction and a concentration through a softplus).
From each patch's sensor query, an MLP gives the device orientation in the body's heading frame as a 6D rotation and the scale of a Laplace distribution of its chordal error.
$\mathcal{L}_{\text{ang}}$ is computed only where the suit's mounting gives a reliable angle label, and $\mathcal{L}_{\text{ori}}$ adds $0.01$ times a penalty on the norms of the two 6D vectors.

\paragraph{Pose head.}
We project the 4\,Hz pose features to 256 channels and add learned part embeddings.
For each part, we average-pool to 2 and 1\,Hz, refine all three rates with residual 1-D convolutions, and fuse from 1 to 2 to 4\,Hz by linear upsampling and addition.
The fused features are then upsampled to 12 and 60\,Hz. At 60\,Hz, we add a skip connection from MLP-embedded raw sensor inputs.
Each part takes a weighted mean over sensors, with weights obtained by attending from a learned part query to the descriptors $\mathbf{d}_i$.
Weights are computed once per window and renormalized over the sensors present at each frame.

A convolution predicts all outputs from each part, retaining only those assigned to that part.
Joint rotations (6D), positions, and uncertainties come from the corresponding part:
torso (pelvis, spine, neck, head), arms (collar, shoulder, elbow, wrist, hand), or legs (hip, knee, ankle, foot).
The torso also predicts root velocity, yaw rate, and pelvis height; each leg predicts its foot contact.

The pose losses in Eq.~(1) of the main paper, with weights in parentheses, are:
squared rotation-matrix error (1);
direct and forward-kinematics joint position errors (2 each);
$L_1$ errors for root velocity (0.5), its 1\,s integral (1), yaw rate (0.2), its 1\,s integral (0.5), and pelvis height (1);
foot-contact binary cross-entropy (0.1);
a 6D norm penalty (0.01);
and Laplace negative log-likelihood of joint position errors under predicted uncertainties (0.05).
Pose and placement ($\lambda_{\text{p}}=0.5$) use separate AdamW moments; their independently normalized updates are summed.

\paragraph{Segment output.}
We use a four-layer Q-Former-style head~\cite{blip2} that attends to sensor and frame tokens from four tapped backbone depths. These eight token sources have separate normalization layers and learned source embeddings, with rotary embeddings encoding patch time. We also include a learned null key.
Each segment has six learned queries positioned at its temporal center, and two additional queries represent window-level context at the window center. Each layer consists of query self-attention, cross-attention, and an MLP.
Self-attention operates over all window queries, masking those of absent segments.
In cross-attention, segment queries attend only to keys within their segment extended by two patches on each side, while context queries attend to all keys. Two MAP probes~\cite{settransformer,scalingvit} independently pool each segment's six output tokens into an alignment vector and a label vector.
Each probe uses a single learned attention query followed by an MLP.
Both vectors pass through a shared normalization layer and linear projection into the 1{,}024-dimensional bge-large embedding space, followed by $\ell_2$ normalization.
The alignment vector is compared with narration embeddings using the SigLIP loss~\cite{siglip} ($\mathcal{L}_{\text{align}}$), and the label vector with embeddings of the 150 activity labels ($\mathcal{L}_{\text{label}}$).
Both losses share a learned temperature and bias.

\paragraph{Caption prompt.}
For one segment, Qwen2.5-0.5B~\cite{qwen25} receives, in its chat format, the text ``Body IMUs:'', one soft token per present sensor (an MLP of $\mathbf{d}_i$), ``Context tokens:'' and the two context tokens, ``Motion tokens:'' and the segment's six tokens (both mapped by one MLP to the LLM width of 896), and the instruction ``Describe C's action in this motion segment, using the contextual motion tokens. Do not invent a location or an object that the evidence does not support.'', where C denotes the wearer as in Nymeria's narrations; the target is the segment's narration.
LoRA~\cite{lora} of rank 16 ($\alpha = 32$, dropout 0.05) adapts the query, key, value and output projections and the three MLP projections of all 24 layers; the LLM's own weights stay frozen.
At inference, captions are decoded greedily with at most 64 new tokens.

\section{Training details}
\label{sec:supp:train}
\paragraph{Input.}
Each sample contains six IMU channels and four magnitudes.
Specific force is expressed in $g$ and clipped to $\pm16\,g$; angular velocity is clipped to $\pm36$\,rad/s and divided by 3.
Both vectors are rotated about the skin normal using the sensor's mean acceleration over valid frames (Sec.~3.2 of the main paper).
The additional channels, $\lVert\mathbf{a}_{xy}\rVert$, $\lVert\mathbf{a}\rVert$, $\lVert\boldsymbol{\omega}_{xy}\rVert$, and $\lVert\boldsymbol{\omega}\rVert$, are invariant to this rotation ($xy$ denotes the skin plane), remaining stable when in-plane gravity is ill-defined.
Each 15-frame patch is flattened to 150 values for the embedding MLP.
Missing frames are zero-filled; tokens with fewer than half their frames valid are masked.

\paragraph{Sensor selection and augmentation.}
The sensor pool comprises 17 suit sensors (measured acceleration and angular velocity derived from fused orientations), glasses, and two wristbands (measured IMUs).
Each window uses a realistic device combination, such as glasses and a watch, with probability $0.3$, or one to six random sensors otherwise.
Mounting augmentation applies a uniform spin in $[-180^\circ,180^\circ]$ about the skin normal, a tilt of up to $15^\circ$ about a random in-plane axis, and a normal flip with probability $0.5$.
We also augment gain, bias, and noise, simulate 20--50\,Hz sampling by box-filtering, decimation, and interpolation back to 60\,Hz, and apply gyroscope dropout indicated by a learned embedding.
For position augmentation, sensors remain at their suit locations ($0.4$), move up to $90^\circ$ around the segment's long axis ($0.4$), or take any angle around that axis ($0.2$), with radii from 3.5\,cm (forearm) to 13\,cm (chest).
Shoulder, hand, and foot sensors are not moved.
\paragraph{Stage 1.}
The decoder, the heads and the loss of Eq.~(1) of the main paper are described in Supplementary~\ref{sec:supp:arch}.
Stage~1 uses windows of 5, 10, 20, or 30\,s and $4 \times 3{,}072$ patches per step.
We train for 25k steps with AdamW ($\beta = (0.9, 0.95)$, learning rate $3 \times 10^{-4}$, 1k warm-up steps, and weight decay $0.05$), taking about 9\,h on four A100s.
The model has 89M parameters, and we use the final checkpoint.
\paragraph{Stage 2.}
We sample windows of one to six consecutive narrated segments.
We merge each pair of adjacent segments with probability 0.15, keeping both narrations and retaining the activity label when the two labels agree.
We crop each segment with probability 0.15 to a random subsegment of at least 2\,s.
We use the losses in Eq.~(2) of the main paper with the output head described in Supplementary~\ref{sec:supp:arch}.
We train 31.8M parameters for 10k steps on four GPUs, taking about 1.6\,h.
We use a learning rate of $3 \times 10^{-4}$ and a LoRA learning rate of $10^{-4}$.
We select the checkpoint on the Nymeria validation split.

\section{Further pose examples}
\label{sec:supp:qual}
\begin{figure*}[tp]
\centering
\begin{subfigure}[t]{0.49\textwidth}
\centering
\includegraphics[width=\linewidth]{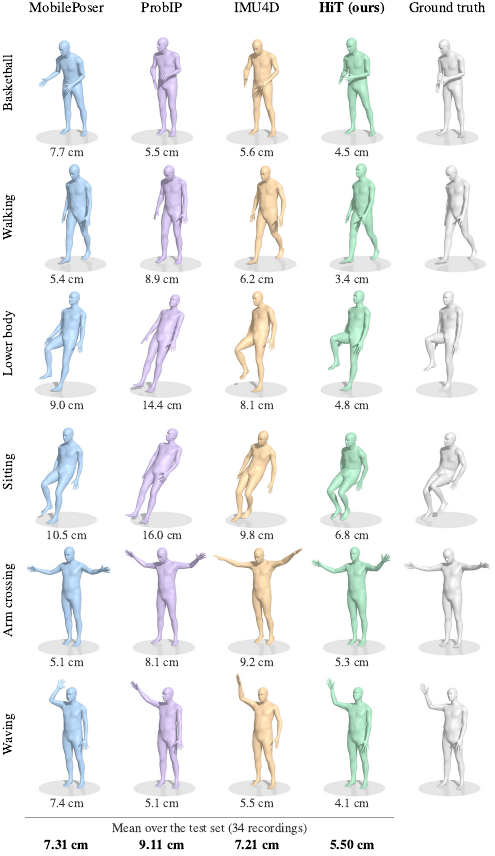}
\caption{Calibrated inputs}
\label{fig:supp_uncal_a}
\end{subfigure}\hfill
\begin{subfigure}[t]{0.49\textwidth}
\centering
\includegraphics[width=\linewidth]{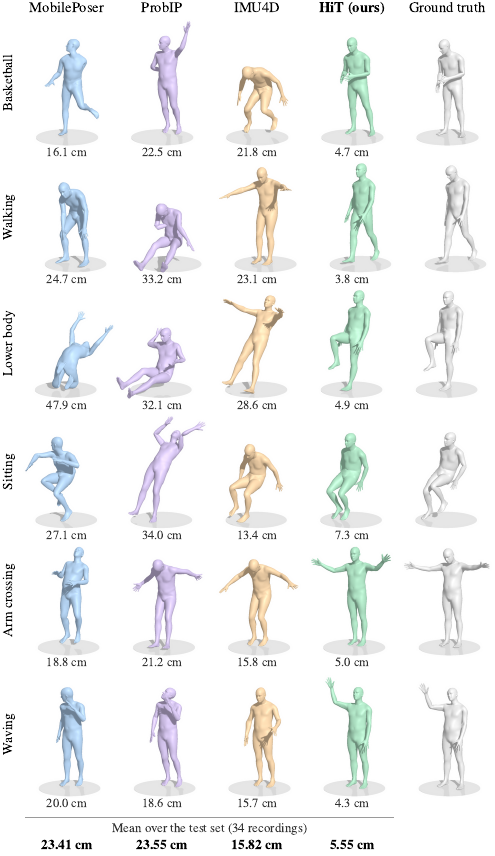}
\caption{Inputs without pose calibration}
\label{fig:supp_uncal_b}
\end{subfigure}
\caption{Six further recordings of IMUPoser (P9--P10) with calibrated inputs (a) and with a random mounting rotation and world heading of every device (b), as in Fig.~4 of the main paper.
Numbers below each pose indicate frame-level MPJPE.
The last line reports mean MPJPE over all 34 test recordings.
Recordings and frames are not picked by hand (Supplementary~\ref{sec:supp:qual}).}
\label{fig:supp_uncal}
\end{figure*}

\begin{figure*}[tp]
\centering
\begin{subfigure}[t]{0.49\textwidth}
\centering
\includegraphics[width=\linewidth]{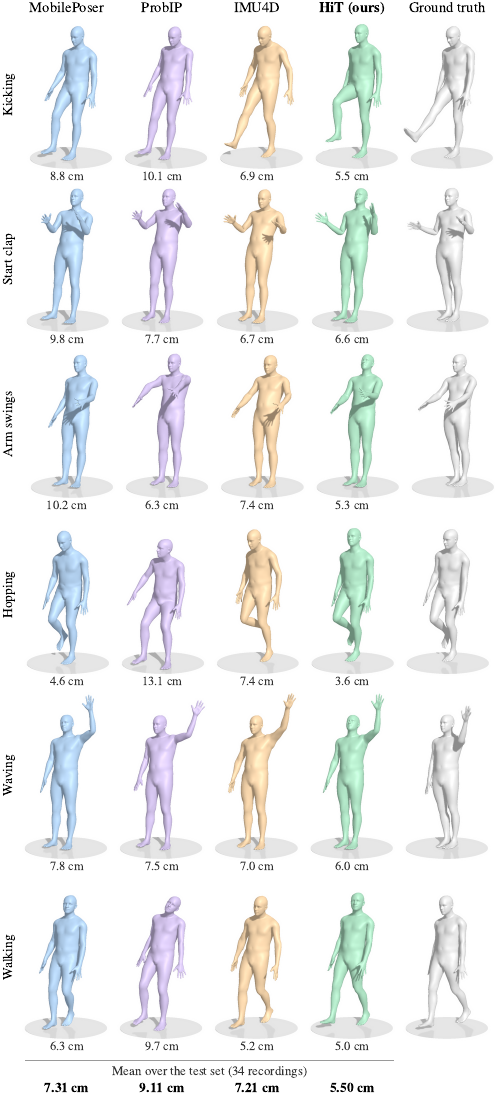}
\caption{Correct placement}
\label{fig:supp_place_a}
\end{subfigure}\hfill
\begin{subfigure}[t]{0.49\textwidth}
\centering
\includegraphics[width=\linewidth]{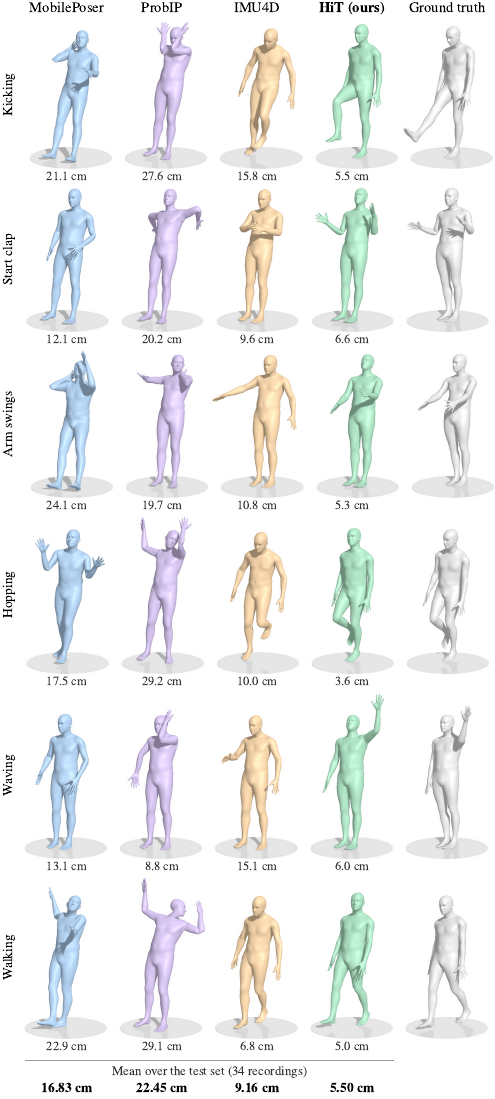}
\caption{Wrong placement, same signals}
\label{fig:supp_place_b}
\end{subfigure}
\caption{Six further recordings of IMUPoser (P9--P10) with correct (a) and incorrect (b) placement labels for the baselines. Both panels use the same calibrated signals.
In (b) MobilePoser and IMU4D are told that the watch is on the right wrist and the phone in the left front pocket, although the signals come from the left watch and the right-pocket phone. For ProbIP, the left and right wrist assignments are swapped.
HiT does not take placement labels as input, so its predictions are identical in both panels.
Numbers below each pose indicate frame-level MPJPE.
The last line reports mean MPJPE over all 34 test recordings.
Recordings and frames are not picked by hand (Supplementary~\ref{sec:supp:qual}).}
\label{fig:supp_place}
\end{figure*}

\Cref{fig:supp_uncal,fig:supp_place} show twelve further recordings of the IMUPoser test participants (P9--P10) in the layout of Fig.~4 of the main paper, with the same models and devices: MobilePoser, IMU4D and HiT read a watch, a pocket phone and the earbuds and are trained on P1--P8; ProbIP reads both watches and the earbuds and starts from the ground-truth pose.

\paragraph{Selection.}
We randomly select 12 test recordings not shown in Fig.~4 using a fixed seed and show six in each figure.
Frames are selected automatically using the same rule as Fig.~4.
Each panel also reports mean MPJPE over all 34 test recordings.

\paragraph{Without calibration.}
\Cref{fig:supp_uncal} repeats the test of Fig.~4 of the main paper: in (b) every device has a random mounting rotation (up to $180^\circ$ about the skin normal and $15^\circ$ of tilt) and a random world heading.
Over the test set this raises the MPJPE of MobilePoser from 7.31 to 23.41\,cm, of ProbIP from 9.11 to 23.55\,cm and of IMU4D from 7.21 to 15.82\,cm, while that of HiT changes from 5.50 to 5.55\,cm.

\paragraph{Wrong placement.}
In \cref{fig:supp_place} all inputs are calibrated, and both panels use the same signals: the left watch, the right-pocket phone and the earbuds.
The baselines receive correct placement labels in (a) and incorrect placement labels in (b).
In (b) MobilePoser and IMU4D are told that the watch is on the right wrist and the phone in the left front pocket. For ProbIP, which reads both watches, the left and right wrist assignments are swapped.
HiT has no input for the placement, so its output in (b) is the same as in (a).
The wrong placement raises the MPJPE of MobilePoser from 7.31 to 16.83\,cm, of ProbIP from 9.11 to 22.45\,cm and of IMU4D from 7.21 to 9.16\,cm, while that of HiT stays at 5.50\,cm.

\end{document}